%% file: main.tex
\documentclass[10pt,twocolumn,letterpaper]{article}

\usepackage[pagenumbers]{iccv} %

\input{preamble}
\definecolor{iccvblue}{rgb}{0.21,0.49,0.74}
\usepackage[pagebackref,breaklinks,colorlinks,allcolors=iccvblue]{hyperref}

\def\paperID{3250} %
\def\confName{ICCV}
\def\confYear{2025}

\usepackage{hyperref}
\usepackage{url}

\input{macros}

\input{math_commands}

\usepackage{graphicx}

\title{Radioactive Watermarks in Diffusion and Autoregressive\\ Image Generative Models}

\author{Michel Meintz$^1$\thanks{Equal Contribution.} \quad Jan Dubiński$^2$$^*$ \quad Franziska Boenisch$^1$ \quad Adam Dziedzic$^1$\\
\noindent \hspace{-0.7em} \normalsize{$^1$CISPA Helmholtz Center for Information Security $^2$Warsaw University of Technology, NASK-National Research Institute}
\\
{\tt\small michel.meintz@cispa.de, jan.dubinski.dokt@pw.edu.pl,\{boenisch,adam.dziedzic\}@cispa.de}
}

\begin{document}
\maketitle
\input{content/00_abstract}

\input{content/01_intro}

\input{content/02_background}

\input{content/03_method}

\input{content/04_problem}

\input{content/05_experiments}

\input{content/06_ablation}

\input{content/07_conclusions}

{
    \small
    \bibliographystyle{ieeenat_fullname}
    \bibliography{main}
}
\appendix
\input{content/08_app_background}
\input{content/09_appendix}

\end{document}

%% file: macros.tex
\usepackage{xparse}
\usepackage{xspace}
\usepackage{tikz}
\usepackage{multirow}
\usepackage{colortbl}
\usepackage{subcaption}

\usepackage{bbold}

\usepackage[textsize=tiny]{todonotes}

\usepackage{amssymb,enumitem}

\newcommand{\mycolspace}{0pt}

\newcommand{\bcode}[1]{\textcolor{blue}{#1}}

\setlist[itemize]{leftmargin=*}
\setlist[enumerate]{leftmargin=*}

\newcommand*{\rej}{{\ooalign{\lower.3ex\hbox{$\sqcup$}\cr\raise.4ex\hbox{$\sqcap$}}}}

\usepackage{stackengine}

\usepackage{xcolor}

\newcommand{\DMs}{DMs\@\xspace}

\newcommand{\ours}{WIAR\xspace} %

\newcommand{\green}{green}
\newcommand{\red}{red}

\graphicspath{{images/}}

\usepackage{booktabs,arydshln}
\makeatletter
\def\adl@drawiv#1#2#3{%
        \hskip.5\tabcolsep
        \xleaders#3{#2.5\@tempdimb #1{1}#2.5\@tempdimb}%
                #2\z@ plus1fil minus1fil\relax
        \hskip.5\tabcolsep}
\newcommand{\cdashlinelr}[1]{%
  \noalign{\vskip\aboverulesep
           \global\let\@dashdrawstore\adl@draw
           \global\let\adl@draw\adl@drawiv}
  \cdashline{#1}
  \noalign{\global\let\adl@draw\@dashdrawstore
           \vskip\belowrulesep}}
\makeatother

\newcommand{\nlp}[1]{}

\newcolumntype{x}[1]{>{\centering\arraybackslash\hspace{0pt}}p{#1}}

\usepackage{amsmath}

%% file: math_commands.tex
\usepackage{amsmath,amsfonts,bm}

\def\eqref#1{equation~\ref{#1}}

\def\1{\bm{1}}

\DeclareMathAlphabet{\mathsfit}{\encodingdefault}{\sfdefault}{m}{sl}
\SetMathAlphabet{\mathsfit}{bold}{\encodingdefault}{\sfdefault}{bx}{n}

\usepackage[ruled,noend,linesnumbered]{algorithm2e}

%% file: content/00_abstract.tex
\begin{abstract}
Image generative models have become increasingly popular, but training them requires large datasets that are costly to collect and curate. To circumvent these costs, some parties may exploit existing models by using the generated images as training data for their own models. In general, watermarking is a valuable tool for detecting unauthorized use of generated images. However, when these images are used to train a new model, watermarking can only enable detection \emph{if} the watermark persists through training and remains identifiable in the outputs of the newly trained model---a property known as \emph{radioactivity}. We analyze the radioactivity of watermarks in images generated by diffusion models (DMs) and image autoregressive models (IARs). We find that existing watermarking methods for DMs fail to retain radioactivity, as watermarks are either erased during encoding into the latent space or lost in the noising-denoising process (during the training in the latent space). Meanwhile, despite IARs having recently surpassed DMs in image generation quality and efficiency, no radioactive watermarking methods have been proposed for them. To overcome this limitation, we propose the first watermarking method tailored for IARs and with radioactivity in mind---drawing inspiration from techniques in large language models (LLMs), which share IARs' autoregressive paradigm. Our extensive experimental evaluation highlights our method's effectiveness in preserving radioactivity within IARs, enabling robust provenance tracking, and preventing unauthorized use of their generated images.

\end{abstract}

%% file: content/01_intro.tex
\section{Introduction}
\label{sec:intro}

\begin{figure*}[htbp]
    \centering
    \includegraphics[width=1\linewidth, trim=0 0 0 0, clip]{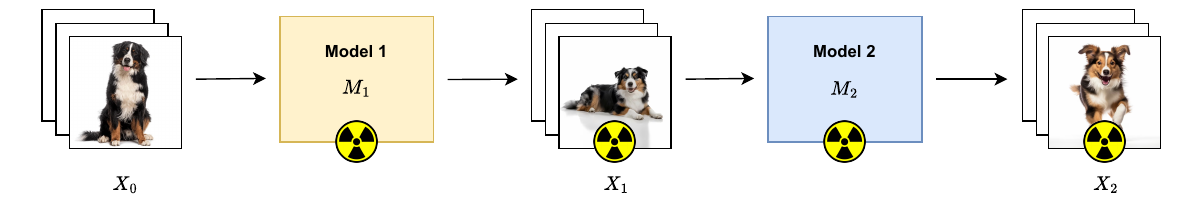} 
    \caption{\textbf{Radioactivity in the image generative models.}
    An image generative model $M_1$ is trained on data $X_0$ and watermarked. Model $M_1$ generates watermarked images $X_1$, which are used to train another image generative model $M_2$. Our goal is to determine whether the watermark signal persists in the images $X_2$ generated by model $M_2$---a property called watermark \textit{radioactivity}.
    }
    \label{fig:teaser}
\end{figure*}

Generative models, particularly in the vision domain~\citep{rombach2022high,ho2020denoising,Song2022improvedScoreBasedGM,tian2024visualVAR,yu2024rar}, have gained significant attention for their ability to generate high-quality, realistic images. However, training these models demands large and diverse datasets, which are often costly and time-consuming to collect and curate~\citep{schuhmann2022laion,oquab2023dinov2,andrews2023ethical}. In light of these high costs, some parties attempt to bypass the data collection process by using images generated by existing models as training data for their own new models~\citep{shumailov2024ai, Sariyildiz_2023_CVPR, tian2023stablerep}. This practice not only undermines the original data sources but also raises critical concerns regarding unauthorized use of generated outputs.

To address these concerns, watermarking has emerged as a valuable tool to help model owners detect and track the misuse of their generated images~\citep{Fernandez_2023_ICCV, wen2023watermarkDMs, gunn2025undetectable, yang2024gaussian}. By embedding distinctive markers within the generated content, watermarking allows model owners to trace the use and provenance of their generated data. However, while watermarking can effectively detect direct misuse of images, its effectiveness is limited in scenarios where the images are not directly published but instead used as training data for new models. In such cases, traditional watermarking methods can only detect unauthorized use \textit{if} the watermark persists through the training process and remains detectable in the outputs of the newly trained model—a property known as \textit{radioactivity}~\citep{sablayrolles2020radioactive}. 

Radioactivity for \textit{generative} models has, so far, been explored solely in the language domain. \citet{sander2024watermarking} demonstrated that when watermarked text is used to fine-tune a new generative large language model (LLM), traces of the original watermark persist in the outputs of the fine-tuned model, enabling detection of unauthorized use. However, similar explorations in the vision domain are lacking.
In this work, we aim to bridge this gap by analyzing the radioactivity of state-of-the-art (SOTA) watermarking methods for image generative models. We depict the problem setup in \Cref{fig:teaser}.

We empirically observe that existing watermarking approaches for diffusion models (DMs), particularly for latent diffusion models (LDMs), fail to exhibit radioactivity, as the watermark is erased either during encoding into the latent space or through the inherent noising-denoising process of training in the latent space.
We also demonstrate that watermarks can transfer when DMs operate directly in the image (pixel) space, highlighting that the latent space is the primary reason for the watermark loss. However, image-space DMs are typically trained on low resolutions (\eg $32\times32$) images and produce significantly lower-quality generations than LDMs, making this approach impractical for high-quality image synthesis.

Next, we turn our attention to image autoregressive models (IARs)~\citep{tian2024visualVAR,yu2024rar}, which have recently achieved new SOTA results in image generation quality and efficiency, surpassing DMs. To date, no radioactive watermarking schemes have been proposed for these models. Motivated by the success of watermarking for LLMs~\citep{kirchenbauer2023watermarkLLM}, which share the autoregressive framework with IARs, we introduce \textbf{\ours}, the first watermarking scheme designed specifically for IARs and with radioactivity in mind.
\ours embeds and detects watermarks in IARs based on their token distribution.
Our extensive experimental evaluation demonstrates that \ours preserves radioactivity, facilitating robust detection of unauthorized use of generated images in IARs. 

In summary, we make the following contributions:
\begin{itemize}
    \item We present the first study on watermark radioactivity for image generative models and show that SOTA watermarking schemes for LDMs lack radioactivity, preventing detection of unauthorized use.
    \item We propose the first watermarking scheme for IARs, ensuring radioactivity by transferring to new models. Our \ours method requires no additional training, embeds at inference time into pre-trained models, and preserves high image quality. 
    \item We conduct a comprehensive empirical evaluation of radioactivity across four SOTA generative models, spanning both diffusion and autoregressive paradigms, and employing five different watermarking methods. 
\end{itemize}

%% file: content/02_background.tex
\section{Background}
\label{sec:background}

We consider the current state-of-the-art image generative methods, including diffusion and autoregressive models, and the corresponding watermarking techniques. 

\subsection{Image Generative Models}
\textbf{Diffusion Models (DMs)}~\cite{ho2020denoising,Song2022improvedScoreBasedGM} are trained by progressively adding noise to the data and then learning to reverse this process. The forward diffusion process adds Gaussian noise $\epsilon \sim \mathcal{N}(0, I)$ to a clean image $x$, yielding a noisy sample $x_t \gets \sqrt{\alpha_t} x + \sqrt{1 - \alpha_t} \epsilon$, where $t\in[0,T]$ is the diffusion timestep, and $\alpha_t \in [0,1]$ is a decaying parameter with $\alpha_0 = 1$ and $\alpha_T = 0$. The model $f_{\theta}$ is trained to predict the noise $\epsilon$ by minimizing: $\mathcal{L}(x, t, \epsilon; f_{\theta}) = \lVert \epsilon - f_{\theta}(x_t, t) \rVert_2^2\text{.}$ In conditional settings, $f_{\theta}$ is guided by an additional input $y$, such as a class label~\cite{ho2020denoising} or a text embedding from a pretrained encoder like CLIP~\cite{radford2021learning}. 
\textbf{Latent diffusion models (LDMs)}~\citep{rombach2022high} improve \DMs by conducting the diffusion process in the latent space instead of in the pixel space, which significantly reduces computational complexity, making training scalable and inference more efficient. For the LDMs, the encoder $\mathcal{E}$ transforms the input $x$ to the latent representation $z=\mathcal{E}(x)$ and the diffusion loss is formulated as $\mathcal{L}(z, t, \epsilon; f_{\theta}) = \lVert \epsilon - f_{\theta}(z_t, t) \rVert_2^2.$
\textbf{Elucidated diffusion models (EDMs)}~\citep{karras2022elucidating_edm}, in contrast, operate in the pixel space (directly on $x$ instead of on its latent representation $z$). 

\textbf{Image Autoregressive Models (IARs).} 
Following the standard next-token-prediction paradigm, for a given set of already predicted tokens $x_{i-1},x_{i-2},...,x_{2},x_{1}$, IARs autoregressively predict the next token $x_{i}$. The autoregressive model $p_{\theta}$ is trained to maximize the probability $p_{\theta}(x_{i}|x_{i-1},x_{i-2},...,x_{2},x_{1})$. Early IARs applied row-by-row raster-scan, z-curve, or spiral orders~\citep{chen2020generatePixels,oord2016pixelcnn,yu2022vectorquantized,esser2021taming}. Recent autoregressive approaches for image generation propose improved self-supervised learning objectives. Visual autoregressive models (VARs) use the next scale (or resolution) prediction as the pretext task~\citep{tian2024visualVAR} while RARs~\citep{yu2024rar} randomly permute the tokens and then perform the standard next token prediction, given the token positional information.
Overall, the shift in the design paradigm of autoregressive models for images, such as VARs and RARs, allowed them to outperform DMs in the image generation quality and efficiency.  

\subsection{Watermarking Techniques}\label{sec:learning_watermark}

\textbf{Watermarking Diffusion Models.}
SOTA watermarking techniques for DMs are learning-based~\citep{zhao2023recipe,Fernandez_2023_ICCV,wen2023watermarkDMs,gunn2025undetectable}. These watermarking methods have three key components: a watermark $(w)$, an encoder $(E)$ and a decoder $(D)$. The encoder receives an input image $X$ and the watermark $w$ as input and is trained to embed the watermark into the image, while not reducing the image quality. Simultaneously, the decoder is trained to reconstruct the watermark given the watermarked image. This can be formulated as: $w = D(E(X, w))$ and $X \simeq X_w = E(X, w) $. Watermarking methods for LDMs, can target different points of the diffusion process, such as the initial noise prediction~\citep{wen2023watermarkDMs,gunn2025undetectable} or the decoding process~\citep{Fernandez_2023_ICCV}. In this work, we consider the following watermarking methods for DMs:

\textit{Recipe Method.} \citet{zhao2023recipe} proposed a method that utilizes a pretrained encoder-decoder structure to embed a predefined binary signature into the training data of an EDM. The EDM is then trained on this watermarked dataset, causing it to reproduce the watermark in its generated images. A watermark decoder is subsequently used to extract and verify the binary signature from these outputs. We refer to this approach to as the \textit{Recipe} method.

\textit{Stable Signature.} \citet{Fernandez_2023_ICCV} introduced a watermarking technique for LDMs~\citep{rombach2022high}, leveraging the model's structure. The method fine-tunes the latent decoder to embed an invisible binary signature in every generated image. This signature can later be recovered using a pretrained watermark extractor.

\textit{Tree-Ring.} \citet{wen2023watermarkDMs} proposed embedding a structured pattern into the \textit{initial noise vector} used in DM sampling. By structuring these patterns in the Fourier space, the method achieves a high level of robustness against common image transformations, such as cropping, dilation, and rotation. Unlike the \textit{Stable Signature} approach, which modifies the decoder, \textit{Tree-Ring} influences the sampling process itself, allowing for watermark detection by inverting the diffusion process and analyzing the retrieved noise vector.

\textit{PRC Watermarking.} \citet{gunn2025undetectable} introduced \textit{P}seudo \textit{R}andom \textit{C}ode (PRC) watermarking, which  utilizes the random generation of an \textit{initial noise vector}. This noise vector is sampled based on a pseudorandom error-correcting code~\citep{Christ2024Pseudorandom}, enabling watermark detection. Given an image, the original noise vector can be reconstructed, allowing the identification of the specific random code used to generate the image.

\textbf{Watermarking Autoregressive Models.} Recently, the main efforts in the area of watermarks for autoregressive models were directed towards designing watermarks for LLMs~\citep{kirchenbauer2023watermarkLLM,christ2024watermarkLLM}. 
We focus on the watermarking method proposed by~\citet{kirchenbauer2023watermarkLLM} due to its simplicity, practicality, and performance. Their technique first randomly divides the set of all possible tokens into a $\green$ list and a $\red$ list, followed by \textit{softly} nudging the generation towards sampling the tokens from the $\green$ list. The detection of the watermark does not require the access to the LLM and is based on how many times the tokens from the $\green$ list instead of the $\red$ list are used in a given text. Making use of the concepts from the LLM watermarking, we propose the first watermarking scheme for IARs.

%% file: content/03_method.tex
\section{Watermarking Method for IARs}
IARs, like LLMs, follow an autoregressive generation paradigm. This suggests that watermarking in IARs should also be based on their generated tokens. However, fundamental differences between IARs and LLMs make direct adaptations of LLM watermarking techniques infeasible.
The key distinctions lie in the structure of their tokens: LLMs operate on text tokens with a \textit{well-defined linear order}, whereas IARs generate image tokens that are spatially arranged without naturally forming a one-dimensional (1D) ordered sequence. Therefore, designing a watermarking scheme for IARs requires first defining a meaningful ordering of tokens. We do so by following the respective inherent generation properties of different IAR types.

Next, LLM tokens are \textit{discrete} and correspond to \textit{discrete} words or subwords, which are directly output without any modifications. The mapping between tokens and text is one-to-one. In contrast, IARs generate tokens that represent \textit{continuous} signals, which are subsequently decoded into images. However, reversing this process by encoding the images and then tokenizing them can result in a different set of tokens as we analyze more in detail in \Cref{sec:ablations}. This discrepancy arises from the imperfect image encoding and decoding. 
Consequently, this introduces a major challenge for watermarking in IARs, where the watermark detection on the level of tokens is inherently more difficult in practice than in LLMs and requires, hence, a robust approach. 
To address this, we identify token sequences with the highest overlap and incorporate them into the watermark detection process.
In the following sections, we detail our proposed \ours watermarking method, including the notation, watermark encoding, and decoding.

\paragraph{Notation.} We denote by $u_i$ the autoregressive unit generated in each step. For example, for RAR, this would be a single token id $x_j$, whereas for VAR, this would be all the tokens output for a given resolution, \ie, $(x_1,\dots,x_{t_i})$, where $t_i$ is the number of tokens for the resolution $i$.

The notation of \Cref{alg:embed} and \Cref{alg:detect} follows \citet{tian2024visualVAR} with the following operations:
\begin{itemize}
    \item \textit{Interpolate}: Up- or down- scaling to the respective resolution.
    \item \textit{Partition}: Split the vocabulary into a green and red list given a random number and size of the green list.
    \item \textit{Bias}: Bias all logits that are part of the green list with the delta value, \ie adding it to the logits of all green list tokens.
    \item \textit{Lookup}: Lookup the tokens in the codebook V and return the quantized values.
\end{itemize}

\subsection{Watermark Embedding} Next, we present our \ours approach for embedding a watermark into IARs. 
Intuitively, \ours relies on $\red$ and $\green$ lists, following~\citep{kirchenbauer2023watermarkLLM} and we nudge the model to generate tokens from the $\green$ list (and not from the $\red$ list) during inference.
Most importantly, our watermark does not rely on any modifications of the model's training procedure and does not require costly retraining or fine-tuning.
We present the details in \Cref{alg:embed}, where we color-code the \bcode{lines for watermarking in blue} and outline the surrounding general image generation steps in a unified manner across different types of IARs. The concrete instantiation of the generation process can be adjusted based on the concrete IAR type. For example, for VARs, the number of tokens per resolution $i$ in lines 6, 10, and 14 varies. In contrast, for RARs, only one token is generated at each step. These differences do not affect the general watermarking procedure.

\paragraph{Embedding a Watermark.} The watermarking starts with hashing the integer representations of the tokens $u_{i-1}$ from the previous $(i-1)$ resolution (line 7). We use the obtained hash value to seed a PRG (Pseudo-Random Generator) in line 8. Using the obtained random seed (rand), in line 9, we randomly partition the vocabulary $V$ (codebook) into a green list $\green$ and a red list $\red$.
The dynamic creation of the lists based on previous token representations, rather than relying on static lists, enables higher-quality generations.
The parameter $\gamma \in (0,1)$ is the scaling factor for the size of the $\green$ list. Thus, the size of the lists are $\gamma|V|$ for the $\green$ and $(1-\gamma)|V|$ for the $\red$ one.
The size of the $\green$ list represents a trade-off between watermark strength and output quality, with smaller values of $\gamma$ resulting in a stronger watermark at the expense of generation quality.
In line 10, we generate the new tokens for the resolution $i$.
For the individual logits $l_j^{(k)}$ (where $k$ is index in the logit vector $l_j$) corresponding to tokens from the $\green$ list ($k \in G$), we add the bias term $\delta$ (line 11). We do not add the bias term to the individual logits corresponding to tokens from the $\red$ list ($k\in R$). The softmax function in line 12 can then be defined as:
\begin{equation}
    p_j =  \begin{cases}  \frac{ \exp(l^{(k)}_j+\delta)}{\sum_{i\in R} \exp(l^{(i)}_j)+\sum_{i\in G} \exp(l^{(i)}_j+\delta)}, \quad k\in G\\
  \frac{ \exp(l^{(k)}_j)}{\sum_{i\in R} \exp(l^{(i)}_j)+\sum_{i\in G} \exp(l^{(i)}_j+\delta)}, \quad k\in R. \label{logitboost}
  \end{cases}
\end{equation}
Intuitively, we softly nudge the selection of each new token $x_j$ to be from the $\green$ list based on the biased probability vector $p_j$.

\begin{center}
  \begin{algorithm}[t]
    \caption{\small{~Image Generation for IARs with \ours \bcode{Watermark Embedding}}} \label{alg:embed}
    \small{
    \textbf{Inputs: } autoregressive model $f_{\theta}$ with parameters $\theta$, vocabulary of possible tokens $V$ (codebook), image decoder $\mathcal{D}$\;
    \textbf{Hyperparameters: } steps $K$ (number of resolutions), resolutions $(h_i,w_i)_{i=1}^{K}$, the number of tokens for resolutions $i$ is $t_i=h_i \cdot w_i$, \bcode{the scaling ratio $\gamma \in (0,1)$ for the size of the $\green$ list}, \bcode{constant $\delta$ added to the logits of the $\green$ list tokens}\;
    $e = \text{init}() $\; 
    $u_0= \{\text{initial seed}\}$\;
    \For {$i=1,\dots,K$}
    {
    $(l_1,\dots,l_{t_i}) = f_{\theta}(\text{interpolate}(e, h_i, w_i))$\;

    \bcode{$\text{seed} = \text{hash}(u_{i-1})$\;}
    \bcode{$\text{rand}=PRG(\text{seed})$\;}
    \bcode{$\text{Green}, \text{Red} = \text{Partition}(V, \text{rand}, \gamma)$\;}
    
    \For {$j=1,\dots,t_i$}
    {
      \bcode{$l_{j} = \text{Bias}(\text{Green}, l_j, \delta)$\;}
      \bcode{$p_j=\text{Softmax}(l_{j})$\;}
      $x_{j} = \text{Sample}(p_j)$\;
    }
  
    $u_i = (x_1,\dots,x_{t_{i}})$\;
    $z_i = \text{lookup}(V, u_i)$\;
    $z_i = \text{interpolate}(z_i, h_K, w_K)$\;
    $e = e + \phi_i(z_i)$\;
    }
    $im = \mathcal{D}(e) $\;
    \textbf{Return: } \bcode{watermarked image im\;}}
  \end{algorithm}
  \vspace{-3em}
  \end{center}

\paragraph{Generating the Watermarked Image.} After the token selection, the image generation follows the standard IAR procedure, \ie, we query the vocabulary of tokens $V$ to obtain their representations $z_i$ per each token index $x_j$ (line 15), interpolate to the full resolution $(h_K,w_K)$ (line 16), and project with $\phi_i$ to the embedding space of image encoding. The embeddings are aggregated across all the resolutions. The final embedding is decoded (line 18) to the watermarked image $im$ and returned (line 19).

\subsection{Watermark Detection} 
Our watermark detection then relies on taking a suspect image, \ie, an image where we want to detect whether a watermark was embedded. This suspect image is then encoded into tokens, and we check whether these tokens stem (mainly) from the $\green$ list (and not from the $\red$ list). Intuitively, images that consists (mainly) of tokens from the $\green$ list are marked as watermarks.
To perform detection, we do not require access to the IAR model.

\paragraph{Detecting the Watermark.} \Cref{alg:detect} presents our \ours watermark detection algorithm where we again color-code the \bcode{lines responsible for the watermark} detection in blue. Lines 3 to 7 mainly follow the standard encoding from IARs to obtain $u_i$, which is the integer representation of tokens for the current resolution $i$.
The counter $C$, defined in line 5, acts as an accumulator and denotes the number of times that tokens from the $\green$ list $G$ are selected to represent the input image $im$. Note that the more times the selected tokens come from the $\green$ list $G$, the higher the probability that a watermark was embedded in the image $im$. In line 6, we iterate over the resolutions $i = 1,\dots,K$.
Next, in a similar vein to watermark embedding from \Cref{alg:embed}, we obtain a random seed from the previous autoregressive unit $u_{i-1}$. The intuition is that if we pick the same initial seed during encoding and decoding with the same pseudo random generator, and if the encoded previous autoregressive unit in the encoded image is the same as during the generation process, we will obtain the same seed and be able to divide the vocabulary into the same $\green$ and $\red$ lists. 
The remaining steps follow standard IAR encoding.
Eventually, to detect the watermark, we analyze whether its tokens stem from the $\green$ or $\red$ lists.
We note that the autoregressive model $f_\theta$ is not used for the watermark detection. Given the input image $im$, we only require the access to the image encoder $\mathcal{E}$, the quantizer $\mathcal{Q}$, and the pseudorandom generator $PRG$.

\begin{center}
  \begin{algorithm}
    \caption{\small{~Image Encoding for IARs with \ours \bcode{Watermark Detection}}} \label{alg:detect}
    \small{
    \textbf{Inputs: } raw image $im$, image encoder $\mathcal{E}$, quantizer $\mathcal{Q}$ to the vocabulary $V$ (codebook)\;
    \textbf{Hyperparameters: } steps $K$, resolutions $(h_i,w_i)_{i=1}^{K}$, \bcode{the scaling ratio $\gamma \in (0,1)$ for the size of the   \textit{Green} list}\;
    $e = \mathcal{E}(im)$\;
    $u_0= \{\text{initial seed}\}$\;
    \bcode{$C=0$\;}
    \For {$i=1,\cdots,K$}
    {
    $u_i = \mathcal{Q}(\text{interpolate}(e, h_i, w_i))$\;
    \bcode{$\text{seed} = \text{hash}(u_{i-1})$\;}
    \bcode{$\text{rand}=PRG(\text{seed})$\;}
    \bcode{$\text{Green}, \text{Red} = \text{Partition}(V, \text{rand}, \gamma)$\;}
    \bcode{$C = \text{Count}(u_i, \text{Green})$\;}
    $z_i = \text{lookup}(V, u_i)$\;
    $z_i = \text{interpolate}(z_i, h_K, w_K)$\;
    $e = e - \phi_k(z_k)$\;
    }
    \textbf{Return: } \bcode{\text{StatisticalTest}($\mathcal{H}_0(C))$\;}
    }
  \end{algorithm}
  \vspace{-2em}
\end{center}

\paragraph{Robust Statistical Testing.}
A naive way to detect the watermark would rely on counting occurrences of $\green$ vs. $\red$ list tokens (line 11). However, this approach might not be reliable due to the inherent problem that tokens generated by an IAR during image generation might not entirely match the tokens resulting from encoding the \textit{exact same} generated image afterwards. This effect is caused by imperfections in the encoding and decoding from continuous image tokens to discrete token ids. We quantify this effect in \Cref{sec:ablations} and find that the inherent token mismatch is substantial, making watermark detection in IARs inherently more challenging than for LLMs where each token id has an exact match with a discrete textual token---making encoding the same sequence \textit{lossless} in practice.

However, we find that we can overcome the problem in IARs by relying on robust statistical testing, following the approach by~\citep{kirchenbauer2023watermarkLLM}, used in LLMs to prevent paraphrasing or synonym replacement attacks.
Therefore, we make our 
\Cref{alg:detect} return the output from the statistical test instead. 
The null hypothesis is that $\mathcal{H}_0:$ \textit{The sequence of tokens is generated with no knowledge about the $\red$ and $\green$ lists}. Since the $\red$ and $\green$ lists are selected at random, a non-watermarked image is expected to consists of $\gamma T$ $\green$ and $(1-\gamma)T$ $\red$ tokens, where $T=\sum_{i=1}^{K}t_i$ is the total number of tokens for the image $im$. The watermarked image should consists of a significantly more $\green$ tokens that a non-watermarked image.

More formally, the color ($\green$ or $\red$) of the next token is a random variable $X$ that follows the Bernoulli distribution. For the non-watermarked image, the expectation (mean) is $E[X]=\gamma$ and the Variance $Var[X]=\gamma(1-\gamma)$. The color for all the tokens of image $im$ can be defined as another random variable $Y$ that follows the Binomial distribution with $E[Y]=\gamma T$ and the Variance $Var[X]=\gamma(1-\gamma) T$. The probability $P$ that a non-watermarked image would consists of only the tokens from the $\green$ list is $\gamma^T$, which is extremely small. For example, for an image with the $256 \times 256$ resolution, we obtain $T=680$ tokens and this probability $P$ would be as low as $\gamma^{680}$.
This enables us to detect the watermark by rejecting the null hypothesis $\mathcal{H}_0$. We follow~\citep{kirchenbauer2023watermarkLLM} and compute the z-statistics:
\begin{equation}
    z = (|s|_G - \gamma T) / \sqrt{T\gamma(1-\gamma)},
    \label{eq:z-statistic}
\end{equation}
where $|s|_G$ is the number of the green tokens in the encoded image. We detect the watermark when the $\mathcal{H}_0$ is rejected for $z > \tau$, where $\tau$ is some pre-defined threshold.

%% file: content/04_problem.tex
\section{Radioactive Watermarks}
\label{sec:method}

Having proposed \ours, the first watermarking method for IARs, we are equipped with watermarking techniques for all the SOTA image generative models.
We now formally define the problem of watermark radioactivity in image generative models, discuss its motivation and relevance, and characterize the threat model considered in this work. 

\textbf{Problem Statement.}
Consider an image generative model $M_1$, trained on a dataset $X_0$ (please see \Cref{fig:teaser}), that has been \textit{watermarked} using a method $w$. This model is then used to generate a collection of images $X_1$, which inherently carry the watermark, \ie, an appropriate detection algorithm, specific to $w$, can successfully identify the watermark signal in $X_1$.
Next, $X_1$ is used as a training dataset for a second generative model, $M_2$. Our goal is to determine whether the watermark signal persists in the images $X_2$ generated by $M_2$. If the watermark remains detectable in $X_2$, we consider the watermarking method $w$ to be \textit{radioactive}, indicating that the watermark transfers through generated data.

\textbf{Motivation and Relevance.} Radioactivity in image generative models is essential for ensuring that watermarking remains effective even when generated images are used as training data for new models. Without it, watermarking can only detect direct misuse, such as unauthorized image distribution, but fails when images are integrated into downstream training pipelines---leaving them vulnerable to untraceable and unauthorized reuse. By enabling persistent watermark traces, radioactivity helps creators, dataset curators, and model developers protecting their intellectual property and enables provenance tracking.

\textbf{Threat Model.}
We consider a realistic threat model in which the intermediate images $X_1$ generated by model $M_1$ are neither retained nor available for verifying watermarks in $X_2$. Model $M_1$ may be accessible via a public API, without any assumptions about user accounts or associated metadata, or deployed locally on-premise. Additionally, we assume only black-box access to model $M_2$, meaning we can generate images but have no knowledge of its architecture, logits, or other internal states.

%% file: content/05_experiments.tex
\section{Empirical Evaluation}\label{sec:experimental_evaluation}
We present our setup, main results, and ablations regarding radioactivity and performance of our \ours watermark.

\subsection{Experimental Setup} 

\textbf{Models and Datasets.}
We evaluate the transferability of watermarks across DMs and IARs. Our study includes two types of DMs: the one operating in image (pixel) space and the other in latent space. For the former, we employ an EDM model architecture \citep{karras2022elucidating_edm} and the CIFAR-10 dataset \citep{Krizhevsky2009LearningML}. For the latter, we follow the setup from the previous work on watermarking for LDMs~\citep{wen2023watermarkDMs,gunn2025undetectable,Fernandez_2023_ICCV} and use Stable Diffusion 2.1~\citep{rombach2022high} with the MS COCO 2014~\citep{lin2014microsoft} dataset. 
For IARs, we consider VAR~\citep{tian2024visualVAR} and RAR~\citep{yu2024rar}
as two representative architectures, using the code and models provided in their respective code repositories. We run the IAR experiments using ImageNet.

\textbf{Watermarking Methods.}
We investigate five watermarking methods, including four watermarking approaches for DMs, namely, \textit{Recipe}~\cite{zhao2023recipe}, \textit{Stable Signature}~\cite{Fernandez_2023_ICCV}, \textit{PRC}~\citep{gunn2025undetectable}, and \textit{Tree-Ring}~\citep{wen2023watermarkDMs}, as well as \ours, our new watermark for IARs. 

\textbf{Radioactivity Evaluation.} For the \textit{Recipe} framework on EDMs, as proposed by \cite{zhao2023recipe}, the first model $M_1$ is trained unconditionally on the watermarked images \(X_0\). In accordance with the method, we embed the watermark directly into the training data \(X_0\) of $M_1$. We then generate 50000 images, matching the number of training samples used for $M_1$, to train a second model, $M_2$, using the same hyperparameters.
For \textit{Recipe} on LDMs, we proceed similarly, however, due to the higher image resolution (32x32 in EDMs vs. 512x512 in LDMs), training from scratch is too computationally expensive. To simulate training from scratch we full fine-tune the models. In the case of \textit{Stable Signature}, \textit{PRC}, and \textit{Tree-Ring}, the watermark is directly embedded into $M_1$’s outputs. Consequently, we only fine-tune $M_2$ on the dataset $X_1$ generated by $M_1$, with the MS COCO captions. We follow a similar procedure for IARs. The outputs from $M_1$ are watermarked and then used to train the $M_2$ model. In the case of LDMs and IARs, we also examine an extreme scenario where DM $M_2$ (and $M_1$ in the case of \textit{Recipe}) is fine-tuned on a single image for an exceptionally high (800) number of epochs to \textit{enforce} the transfer of a specific watermark instance.

\textbf{Metrics.} 
We report the \textbf{TPR@FPR=1\%}, which measures the true positive rate (TPR) at a 1\% false positive rate (FPR) when detecting watermarked images. In this case, TPR = 1\% corresponds to random guessing, whereas a significantly higher TPR ($>>1\%$) indicates a robust watermark. 

To assess the detectability of the watermarks, we conduct a statistical test using a \textit{z-test} (see \Cref{eq:z-statistic}). The null hypothesis $\mathcal{H}_0$ states: \textit{"The watermark is not present in the model outputs if the TPR$\le$1\%@FPR=1\%."}
Similarly to the statistical test in \ours, we set $\tau=4$ (following~\citep{kirchenbauer2023watermarkLLM}), meaning that z-values below this threshold indicate a clear detection of the watermark in the outputs from $M_1$ or strong radioactivity in the outputs from $M_2$. The threshold ensures the low probability $3 \times 10^{-5}$ of detecting the watermarks or radioactivity (incorrectly rejecting $\mathcal{H}_0$) when none is present in the image. Conversely, when the z-value is above the threshold $\tau$ 
then the test is inconclusive and most likely the watermark was not present in the image.

\subsection{Analyzing Watermark Radioactivity}
\label{sec:results}

We analyze the watermark radioactivity in \Cref{tab:bitwiseaccuracy_main_results}. The results show that the existing watermarks for EDMs, as well as our new watermark for IARs transfer. In contrast, watermarking schemes for LDMs are not radioactive. 

\renewcommand{\mycolspace}{2pt}
\addtolength{\tabcolsep}{-\mycolspace} 
\begin{table}[h!]
    \caption{\textbf{Watermarks are Radioactive for EDMs and IARs but not for LDMs.} We analyze five different watermarking methods for three different types of image generative models. We report TPR@FPR=1\% for all methods (\ie, 1\% corresponds to the random guessing baseline).}
    \label{tab:bitwiseaccuracy_main_results}
    \centering
    \scriptsize
    \begin{tabular}{ccccccc}\toprule
    \textbf{Type of $M_1$} & \textbf{Type of $M_2$} & \textbf{Method} & \textbf{Output of $M_1$} & \textbf{Output of $M_2$} \\
    \midrule
    EDM & EDM & Recipe &$82.5$ & $82.2$ \\
    LDM & EDM & Tree-Ring &$100$ & $80.39$ \\
    \midrule
    LDM & LDM & Recipe&$0.32$ & - \\
    
    LDM & LDM & Stable Signature & $100$ & $1$ \\
    LDM & LDM & PRC  & $100$ & $1$ \\
    LDM & LDM & Tree-Ring &$100$  & $1$  \\
    \hdashline 
    VAR-\textit{d}16 & VAR-\textit{d}16 & \ours & $94.0$ & $36.1$\\
    VAR-\textit{d}20 & VAR-\textit{d}20 & \ours  & $94.7$ & $36.5$\\
    VAR-\textit{d}24 & VAR-\textit{d}24 & \ours & $97.1$ & $38.1$\\
    VAR-\textit{d}30 & VAR-\textit{d}30 & \ours  & $95.5$ & $37.2$\\
    \hdashline
    RAR-B & RAR-B & \ours & $92$ & $33$ \\ 
    RAR-L & RAR-L & \ours & $90$ & $32$ \\ 
    RAR-XL & RAR-XL & \ours & $94$ &$36$\\ 
    RAR-XXL & RAR-XXL & \ours & $90$ &$29$ \\ 
    \bottomrule
    \end{tabular}
    \vspace{1em}
\end{table}
\addtolength{\tabcolsep}{\mycolspace}

\textbf{Watermarks in EDMs are Radioactive.} 
\label{sec:insights_edm}
For \textit{Recipe} which adds the watermarks directly to the images, the watermark detection rate is $100\%$ in the watermarked original training data \(X_0\). This data is then used to train $M_1$. We find that the watermark successfully transfers from the output of EDM $M_1$ to EDM $M_2$ (see the first row in \Cref{tab:bitwiseaccuracy_main_results}). This aligns with expectations: when a watermark is present in the training data and propagates to the output of $M_1$, it should, by transitivity, implicitly transfer to the output of $M_2$ as well, given that both models perform the same task. The EDM architecture facilitates this transfer, as images are directly generated and learned in the image space. This direct reconstruction of pixel values, without abstraction into the latent space, enhances the retention of watermarks.

\vspace{0.5cm}
\renewcommand{\mycolspace}{2pt}
\addtolength{\tabcolsep}{-\mycolspace} 
\begin{table}[t]
    \centering
    \scriptsize
    \caption{\textbf{Extensive fine-tuning on a single watermarked sample does not enforce watermark transfer in LDMs.} Given the observed lack of watermark transferability in LDMs, we extend our analysis from \Cref{tab:bitwiseaccuracy_main_results} to an extreme scenario aimed at transferring a specific watermark instance. We use the metrics from \Cref{tab:bitwiseaccuracy_main_results}. Additionally, we observe that neither the Tree-Ring watermark transfer from LDMs to RAR-B nor \ours transfers to LDMs.
    \label{tab:watermark_single_tab}
    }
    \begin{tabular}{cccccccc}\toprule
    \textbf{Type of $M_1$} & \textbf{Type of $M_2$}& \textbf{Method} &\textbf{Output of $M_1$} & \textbf{Output of $M_2$} \\
    \midrule
    LDM & LDM & Recipe &$1$ & $1$ \\ 
    LDM & LDM & Stable Signature & $100$ & $0.9$ \\
    LDM & LDM & PRC  & $100$ & $1$ \\
    \hdashline 
    LDM &LDM & Tree-Ring & $100$ & $1$  \\
    LDM & RAR-B & Tree-Ring & $100$ & $1$ \\
    RAR-B & LDM & \ours & $100 $ & $1$ \\
    VAR-\textit{d}16 & LDM & \ours & $100$ & $1$  \\
    \hdashline
    VAR-\textit{d}16 & VAR-\textit{d}16 & \ours & $100 $ & $100$ \\
    VAR-\textit{d}20 & VAR-\textit{d}20 & \ours & $100 $ & $100$ \\
    VAR-\textit{d}24 & VAR-\textit{d}24 & \ours & $100 $ & $100$ \\
    VAR-\textit{d}30 & VAR-\textit{d}30 & \ours & $100 $ & $100$ \\
    \hdashline
    RAR-B & RAR-B & \ours & $100$ & $100$\\
    RAR-L & RAR-L & \ours & $100$ & $100$\\
    RAR-XL & RAR-XL & \ours & $100$ & $100$\\
    RAR-XXL & RAR-XXL & \ours & $100$ & $100$\\

    \hdashline
    VAR-\textit{d}16 & RAR-B & \ours & $100$ & $1$ \\
    RAR-B & VAR-\textit{d}16 & \ours & $100$ & $1$ \\
    \bottomrule
    \end{tabular}
    \vspace{-1em}
\end{table}
\addtolength{\tabcolsep}{\mycolspace} 

\vspace{-0.5cm}
\textbf{Watermarks in LDMs are not Radioactive.}
\Cref{tab:bitwiseaccuracy_main_results} highlights that watermarks for LDM do not transfer, \ie, are not radioactive. 
For example, for Tree-Ring, Stable Signature and PRC, the 1.1\% TPR are close to the 1\% random guessing. For Recipe, the watermark can already not be detected anymore in the output of $M_1$, preventing transfer to $M_2$. 
To study the lack of transferability in the limit, we conduct an additional experiment where we we fine-tune LDMs on a single watermarked instance over 800 epochs to force watermark transfer.  
\Cref{tab:watermark_single_tab} shows that even under such conditions, the watermark does not transfer through fine-tuning for LDMs, resulting in random detection outcomes. 
Next, we analyze the reasons contributing to this outcome.

\begin{figure}[t]
    \centering
    \includegraphics[width=1.0\linewidth]{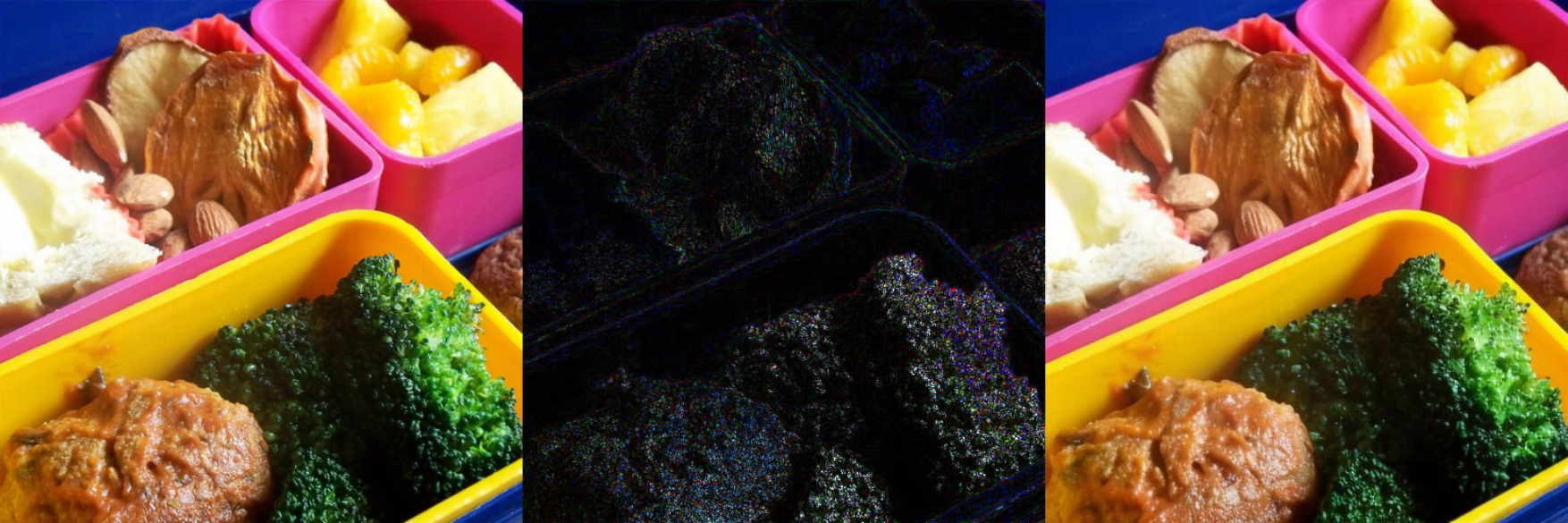}
    \caption{\textbf{The LDM autoencoder acts as a filter for the \textit{Recipe} data-level watermark. }
    \textit{Left}: original watermarked image with $100\%$ watermark detection, \textit{Middle}: difference $\times$ 2, \textit{Right}: image after en-decoding by the SD2.1 autoencoder structure with TPR=1\%@FPR=1\%, \ie, random guessing (meaning that no watermark is detected).
    }
    \label{fig:recipe_vae}
    \vspace{-0.5em}
\end{figure}

\textit{Reason 1: Autoencoder.} 
For the \textit{Recipe} method, we observe that the watermark successfully transfers in EDMs but fails under full fine-tuning in the LDM architecture.  
We attribute this failure to the variational autoencoder (VAE) used in the SD 2.1, which encodes images into the latent space and then back to the image space. In the case of \textit{Recipe}, the autoencoder effectively acts as a filter, suppressing high-frequency details, including embedded watermarks. \Cref{fig:recipe_vae}
presents an example of a watermarked image used for fine-tuning SD 2.1. We observe that, after passing the watermarked images through the VAE, the watermark is no longer detectable.

\begin{figure}[h]
    \centering
        \begin{subfigure}{0.22\textwidth}
        \centering
        \includegraphics[width=0.675\linewidth]{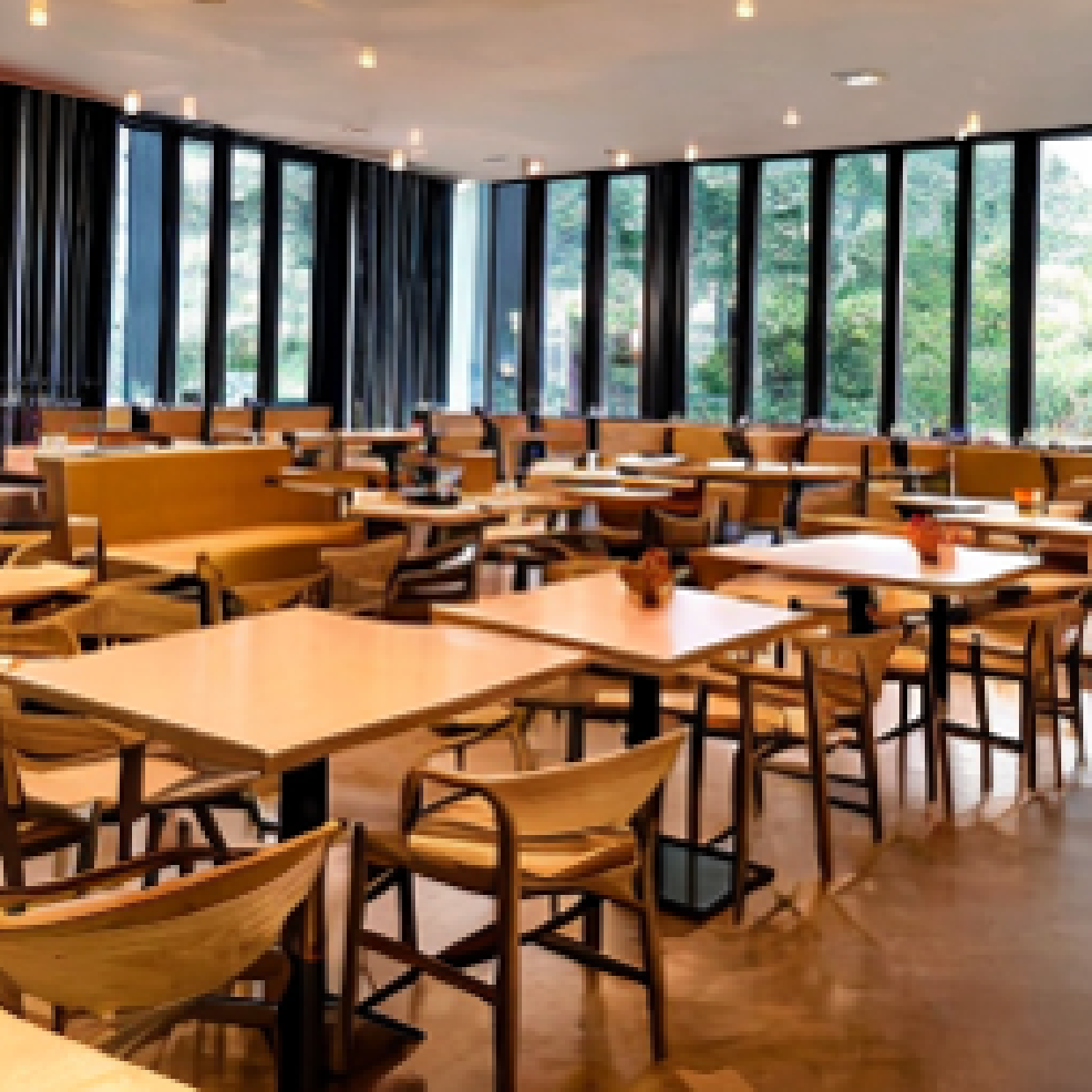}
        \caption{$M_1$: TPR=100\%@FPR=1\%}
        \label{fig:TreeRingsub1-app}
    \end{subfigure}
    \hfill
    \begin{subfigure}{0.22\textwidth}
        \centering
        \includegraphics[width=0.675\linewidth]{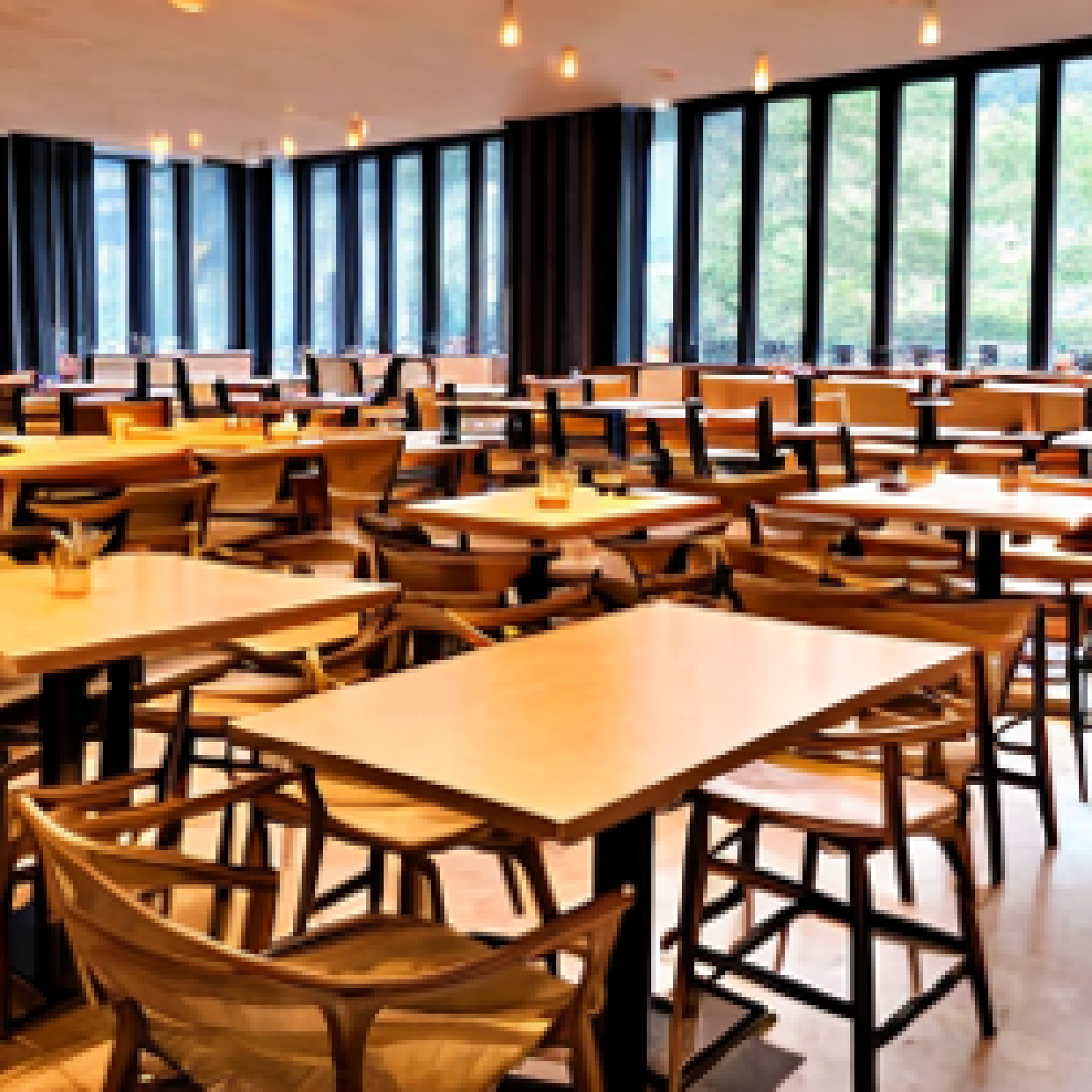}
        \caption{$M_2$: TPR=1\%@FPR=1\%}
        \label{fig:TreeRingsub2-app}
    \end{subfigure}
    \hfill
    \begin{subfigure}{0.22\textwidth}
        \centering
        \includegraphics[width=0.675\linewidth]{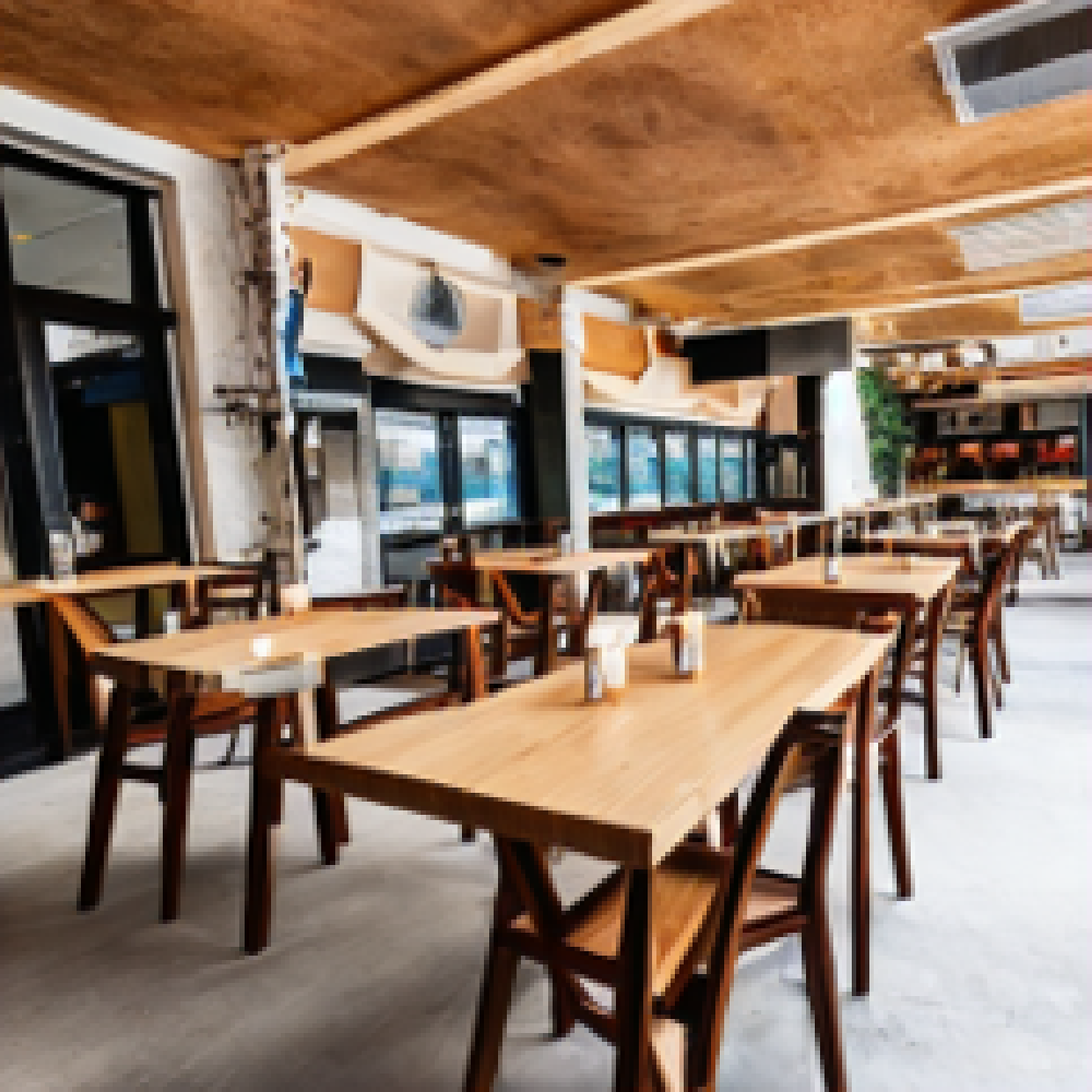}
        \caption{$M_1$: bitwise accuracy $100\%$}
        \label{fig:StableSignature_sub1}
    \end{subfigure}
    \hfill
    \begin{subfigure}{0.22\textwidth}
        \centering
        \includegraphics[width=0.675\linewidth]{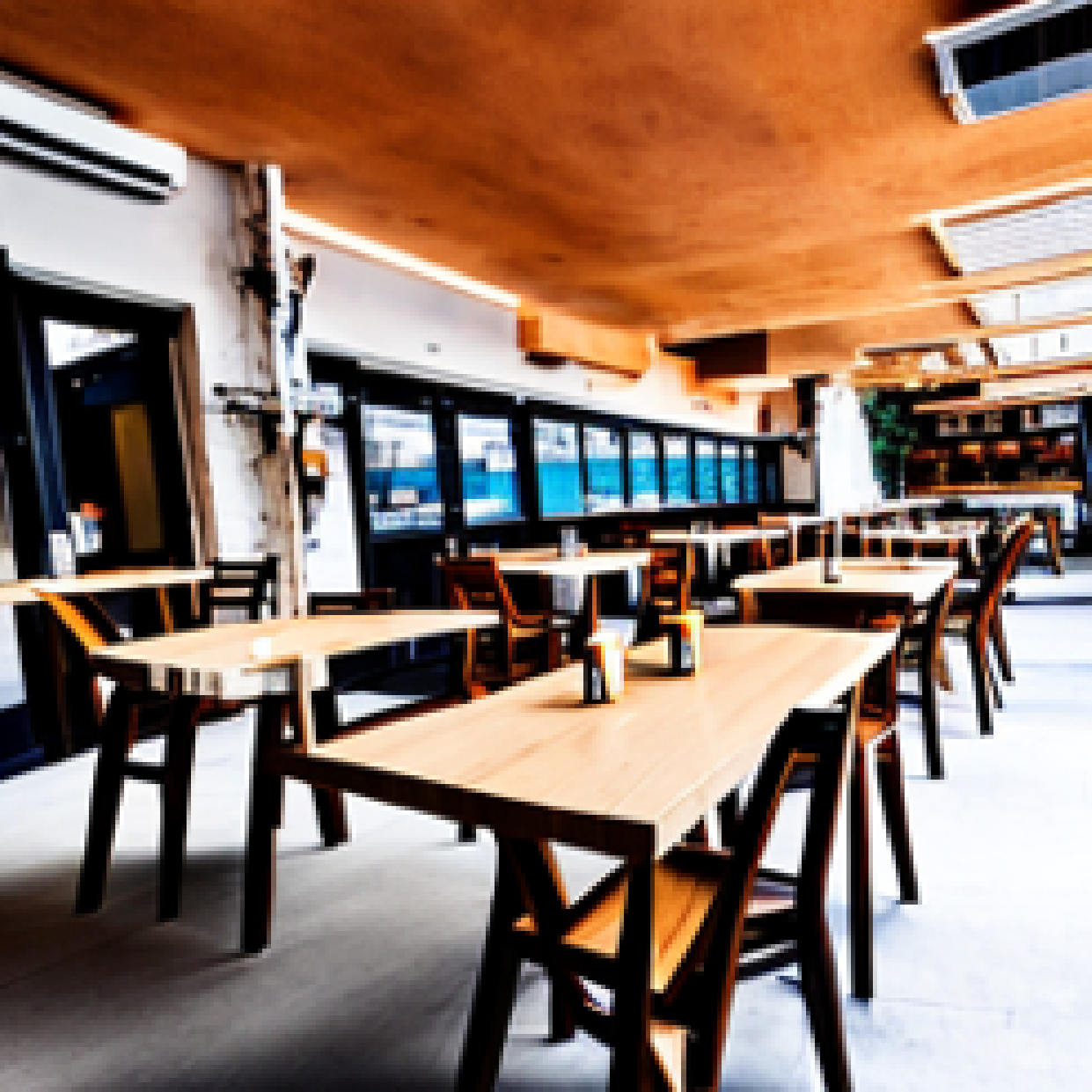}
        \caption{$M_2$: bitwise accuracy $49.9\%$}
        \label{fig:StableSignature_sub2}
    \end{subfigure}
    \hfill
    \begin{subfigure}{0.22\textwidth}
        \centering
        \includegraphics[width=0.675\linewidth]{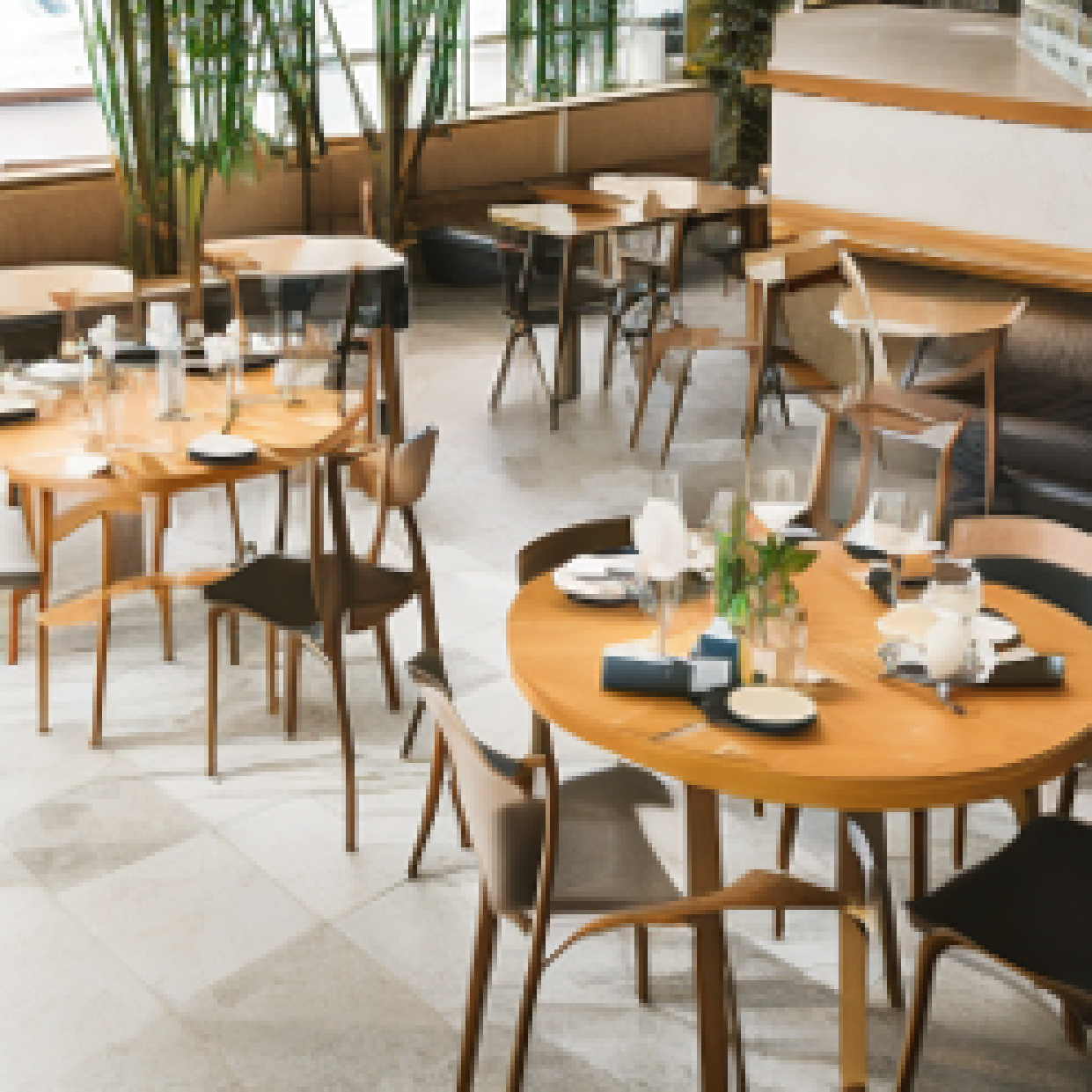}
        \caption{$M_1$: bitwise accuracy $100\%$}
        \label{fig:PRCsub1}
    \end{subfigure}
    \hfill
    \begin{subfigure}{0.22\textwidth}
        \centering
        \includegraphics[width=0.675\linewidth]{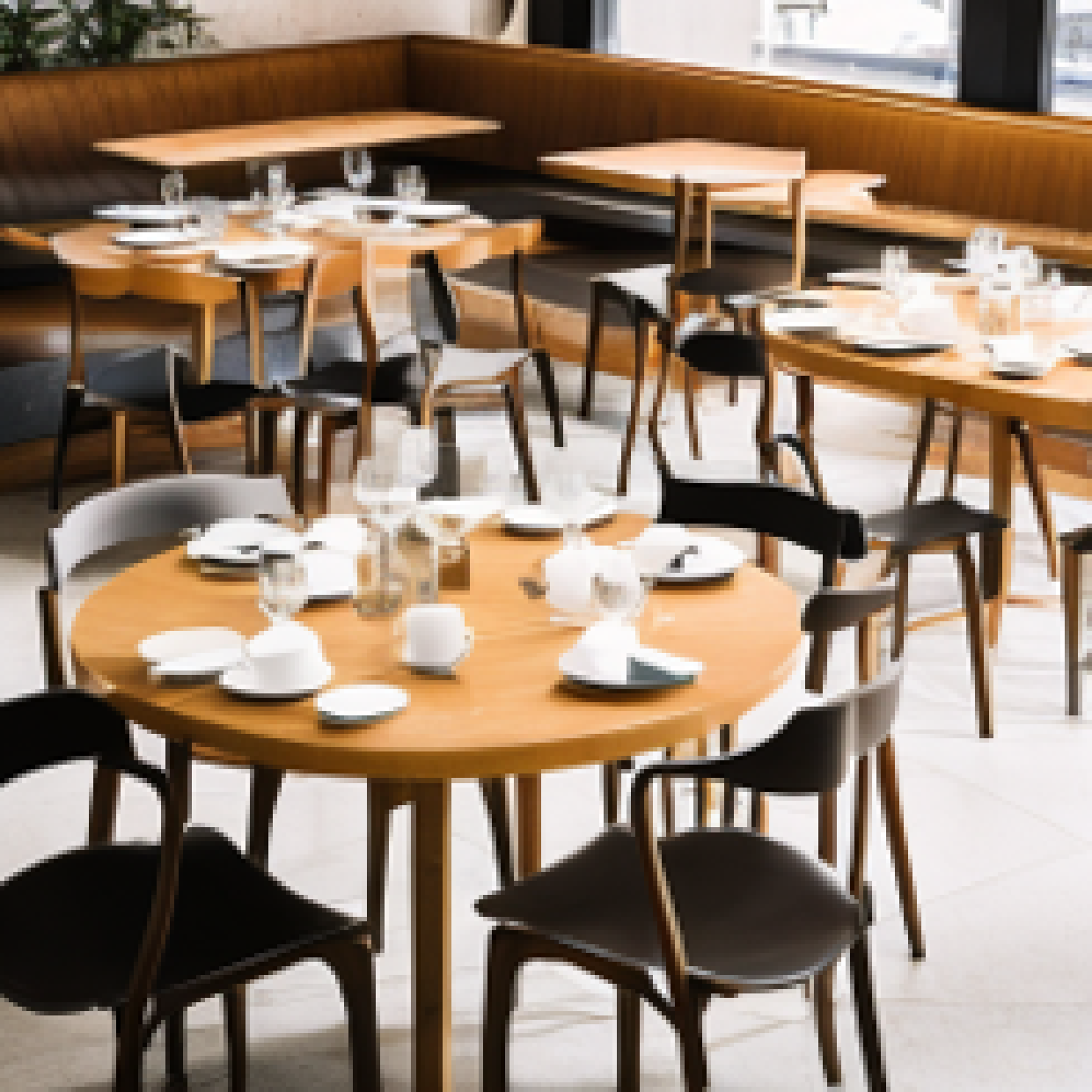}
        \caption{$M_2$: bitwise accuracy $50\%$}
        \label{fig:PRCsub2}
    \end{subfigure}
    \hfill
    \begin{subfigure}{0.22\textwidth}
        \centering
        \includegraphics[width=0.675\linewidth]{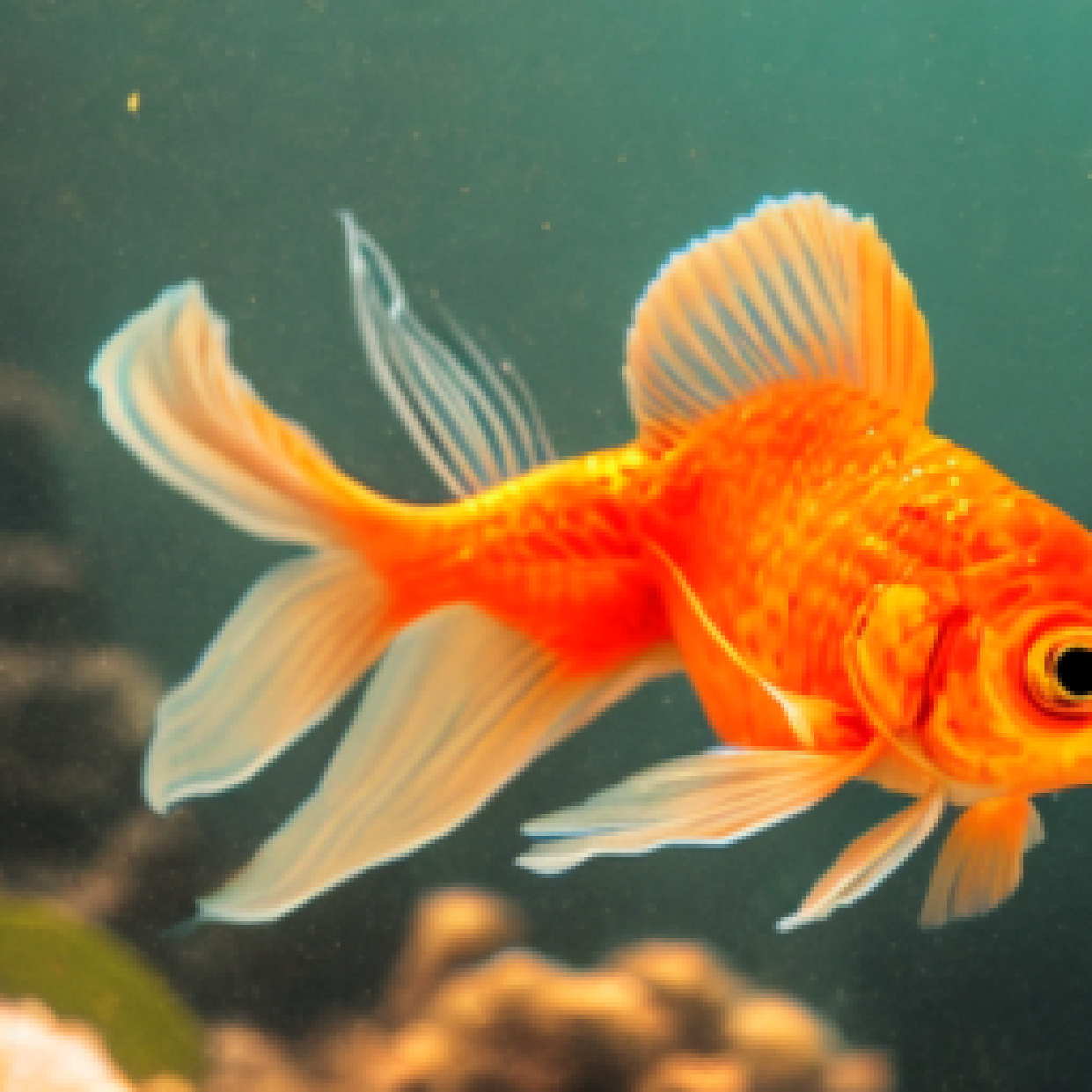}
        \caption{$M_1$: TPR=100\%@FPR=1\%}
        \label{fig:treering_sub1_IAR}
    \end{subfigure}
    \hfill
    \begin{subfigure}{0.22\textwidth}
        \centering
        \includegraphics[width=0.675\linewidth]{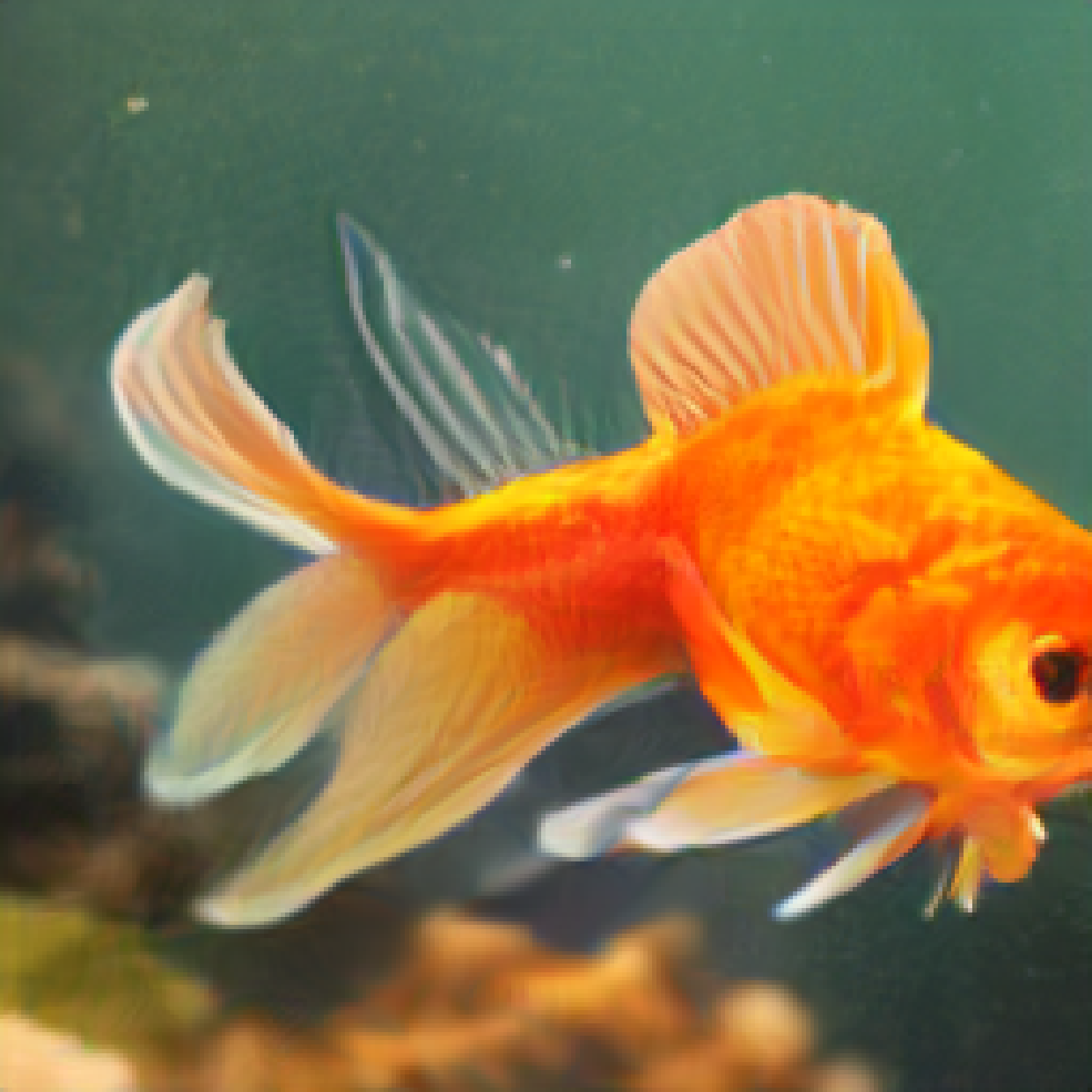}
        \caption{$M_2$: TPR=1\%@FPR=1\% }
        \label{fig:treering_sub2_IAR}
    \end{subfigure}
    \caption{\textbf{The watermarks for LDMs are not radioactive.} We present a qualitative comparison between the outputs from $M_1$ (first column with cases a, c, e, and g) and $M_2$ (second column with cases b, d, f, and h). The first row (a \& b) is for \textit{Tree-Ring} (transfer from LDM to LDM). The second row (c \& d) is for \textit{Stable Signature} (transfer from LDM to LDM), the third row (e \& f) for \textit{PRC} (transfer from LDM to LDM), and the fourth row (g \& h) for \textit{\ours} (transfer from RAR-B to LDM).}
    \label{fig:app_compare_outputs_M1_M2}
\end{figure}

\textit{Reason 2: Noising-Denoising in Latent Space.}
The \textit{Stable Signature}, \textit{PRC}, and \textit{Tree-Ring}  watermarking techniques fail to transfer from LDM $M_1$ to LDM $M_2$. Additionally, \textit{\ours} fails to transfer from RAR-B $M_1$ to LDM $M_2$. However,  as presented in \Cref{tab:after_ae} in the Appendix, we find that for these methods, the watermark remains intact after being encoded into the latent space of the SD 2.1. 
Furthermore, we observe that while \textit{Stable Signature}, \textit{PRC}, and \textit{\ours} are erased during the down-sampling to the low $32\times32$ resolution used in the training of EDMs, the Tree-Ring watermark persists the down-sampling and the \textit{noising-denoising process \textbf{in the pixel space}} while training EDM $M_2$ (see the transfer from LDM $M_1$ to EDM $M_2$ for Tree-Ring in \Cref{tab:bitwiseaccuracy_main_results}). This suggests that the Tree-Ring watermark is erased during the \textit{noising-denoising process \textbf{in the latent space}} while training LDM $M_2$. CtrlRegen~\cite{liu2025image_ctrlregen} is an attack based on this process, which can remove watermarks such as Tree-Ring and StableSignature.
We present examples of images generated by $M_1$ and $M_2$ with the \textit{Tree-Ring} in \Cref{fig:app_compare_outputs_M1_M2}. The watermarks are easily detectable in the outputs from LDM $M_1$, whereas they are no longer traceable in the outputs from LDM $M_2$. Additionally, the images from $M_1$ vs $M_2$ exhibit noticeable differences in composition and color. While the semantic structure remains consistent, the spatial layout differs significantly. This variation in image composition resembles a \textit{visual paraphrasing attack}~\citep{barman2024brittlenessaigeneratedimagewatermarking}. 

\textbf{Our Watermark for IARs is Radioactive.}
We observe that our new watermarks for IARs successfully transfer from $M_1$ to the subsequently trained $M_2$ IARs. In the extreme scenario, see \Cref{tab:watermark_single_tab}, where we fine-tune $M_2$ on a single image, the watermark exhibits even \textit{perfect} radioactivity (100\% transfer). %
In the practical scenario, when fine-tuning IARs on 40k images, in \Cref{tab:bitwiseaccuracy_main_results}, we observethat the TPR at 1\% FPR is around 30-40\%, \ie significantly higher than random guessing (1\%). Thereby, our \ours watermark remains the only watermark that consistently transfers between models of high-generation quality.

\begin{figure}[t]
    \centering
    \includegraphics[width=1.0\linewidth]{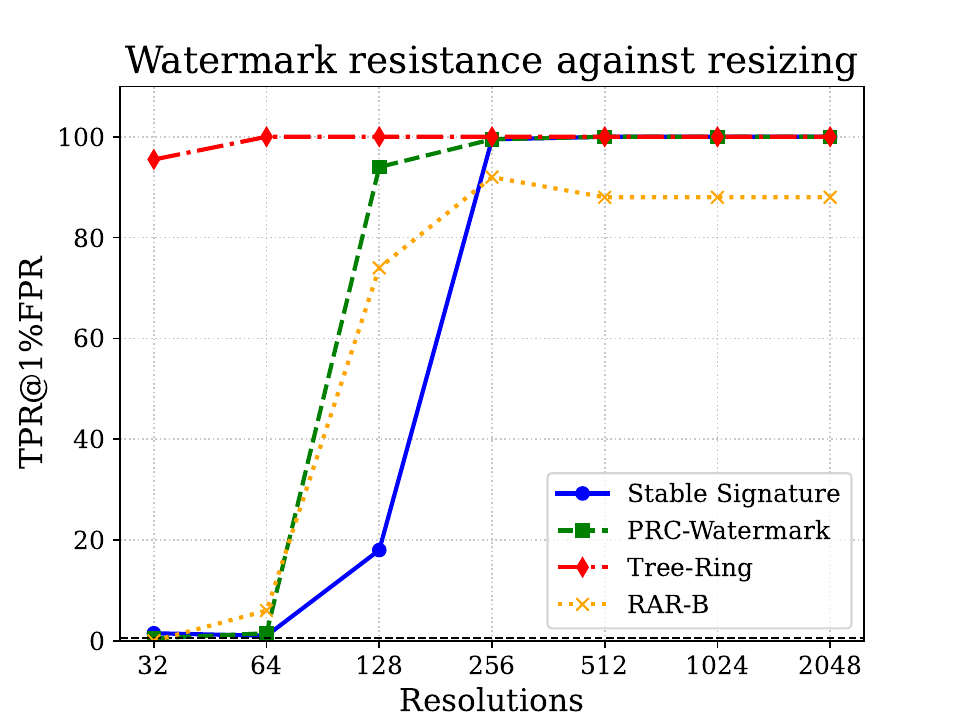}
    \caption{\textbf{Down-sampling removes all the watermarks apart from Tree-Ring that transfers to EDMs.} We analyze the strength of the different watermarks under strong resizing. We observe that only Tree-Ring, which, in contrast to other methods, does not operate in the image but in the frequency space, is resilient to the down-sampling in the pixel space. We use the Gustavosta Stable Diffusion Prompts, usually used to evaluate image generation quality, to generate 200 images at a starting resolution of $512\times512$, which we resize to the target resolution. For detection, the images are reverted back to $512\times512$. 
    }
    \label{fig:resizing_resitance}
    \vspace{-1em}
\end{figure}

\textbf{Transferability between Architectures.}
In line with our threat model, we also analyze watermark transferability across different model architectures. Specifically, we evaluate transferability to EDMs, which generate lower-resolution images. Since high-resolution outputs from other generative models must be downscaled to $32\times32$ pixels, this introduces an additional challenge for watermark retention.
As shown in \Cref{fig:resizing_resitance}, only the Tree-Ring watermark successfully transfers to EDMs. Unlike other methods, Tree-Ring is embedded in the Fourier space rather than the image space, making it robust to various augmentations such as convolutions, cropping, dilation, flipping, rotation, and even resizing. In contrast, downsampling alone eliminates \ours, as well as the \textit{Stable Signature} and \textit{PRC} watermarks, erasing any transferable signal.
While resilience to downsampling is generally desirable, reducing image resolution from $256\times256$ or $512\times512$ to $32\times32$ also significantly degrades visual quality---reducing the advantages of utilizing the generated data in the first place. %

We also analyze transferability of watermarks \textit{from} and \textit{to} LDMs in \Cref{tab:watermark_single_tab}. We find that the watermark from RAR-B does not transfer to LDMs. This can be attributed to the training process of the LDM, which, as previously explored, removes the watermarks. For the transfer from LDM to RAR-B with the Tree-Ring watermark, we find that the tokenizer in IAR models acts as a filter against the watermark and removes it.

%% file: content/06_ablation.tex
\subsection{Ablations for Watermarking in IARs }
\label{sec:ablations}

\begin{figure}[t]
    \centering
    \begin{subfigure}[b]{0.45\textwidth}
        \centering
        \includegraphics[width=0.82\linewidth]{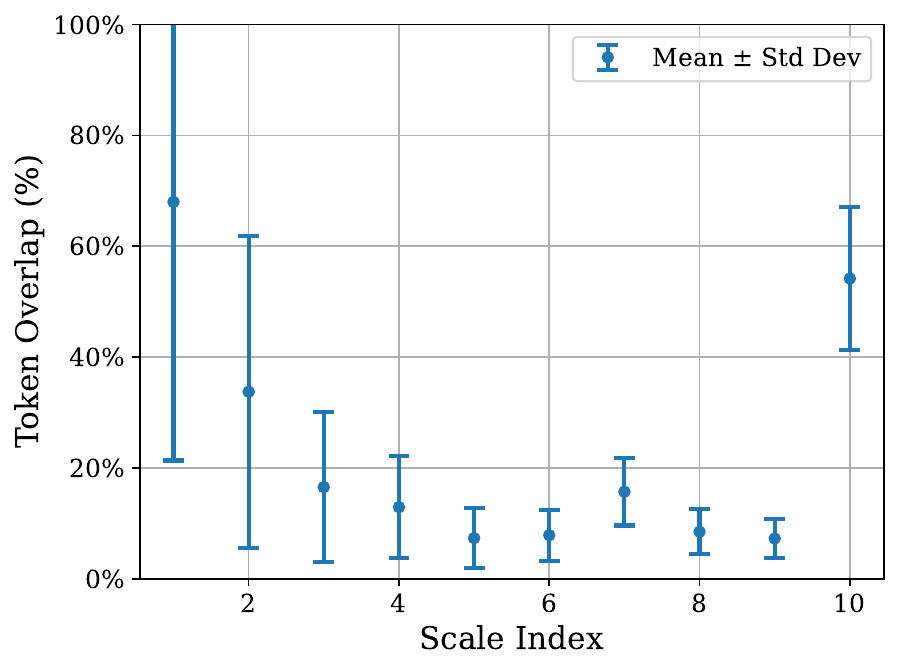}
        \caption{VAR.}
        \label{fig:var_overlap}
    \end{subfigure}
    \quad
    \begin{subfigure}[b]{0.45\textwidth}
        \centering
        \scriptsize
        \vspace{2em}
            \begin{tabular}{cc}\toprule
     Type of RAR & Avg. Token Overlap \\
     \midrule
    RAR-B & $65.45\% \pm 9.04\% $ \\
    RAR-L & $67.62\% \pm 7.21\% $ \\
    RAR-XL & $65.75\% \pm 6.38\% $ \\
    RAR-XXL & $65.68\% \pm 8.02\% $ \\
    \bottomrule
    \end{tabular}
        \caption{RAR.}
        \label{fig:rar_overlap}
    \end{subfigure}
    \caption{\textbf{Token overlap between generated vs encoded images for VAR-\textit{d}30 and RAR.} Tokens from decoding during the image generation differ from tokens obtained during image encoding.}
    \label{fig:token_overlap}
    \vspace{0.3em}
\end{figure}

\textbf{Token Mismatch in Decoded (Generated) vs. Encoded Images.} We identify an inherent challenge in IAR watermarking arising from imperfect image encoding and decoding. Specifically, a model generates tokens $T$, which are then decoded into a generated image. When \textit{the exact same image} is subsequently encoded and tokenized, it produces a different set of tokens, $T'$, which do not fully match the original tokens $T$, \ie $T'\ne T$. This differs from LLMs, where the tokens decoded to the generated \textit{text} match the tokens obtained from the tokenized exact same \textit{text}. The discrepancy decreases the watermark detection rate of \ours, as it relies on the correct tokens to determine whether they belong to the $\green$ or $\red$ list. We quantify the token overlap in \Cref{fig:token_overlap} for VAR and RAR. 
For VAR, the initial and final scales (resolutions) exhibit the highest token overlap. This is likely because the initial resolutions capture broad structural features, while at the final resolution, the image is already refined, resulting in fewer token changes.
For RAR, which only generates one token at a time, we observe that the average token overlap is similar among all model sizes, namely around 65\%.

\textbf{Watermarking vs. Generation Quality.} We analyze the impact of our \ours on the quality of generated images (measured in FID score~\citep{heusel2017gans}) in \Cref{tab:watermark-image-quality}.
The results show that our \ours only marginally affects image quality, highlighting that our method is able to provide robust provenance tracking without sacrificing generative performance.

\renewcommand{\mycolspace}{0pt}
\addtolength{\tabcolsep}{-\mycolspace} 
\begin{table}[t]
    \caption{
    \textbf{Impact of watermarking on image generation quality.} We measure the performance difference in the generation quality between \textit{clean} non-watermarked models vs. the generation with the \textit{watermark} embedding. We observe a slight degradation in image quality, however, the absolute image generation quality is still high for IARs watermarked with \ours. %
    }
    \vspace{0.5em}
    \label{tab:watermark-image-quality}
    \centering
    \scriptsize
    \begin{tabular}{cccc}\toprule
    \textbf{Model} & \textbf{\# of Parameters} & \textbf{Clean FID}$\downarrow$ & \textbf{Watermark FID}$\downarrow$\\
    \midrule
    VAR-\textit{d}16 & 310M & 6.50 & 7.41\\
    VAR-\textit{d}20 & 600M & 5.77 & 6.63 \\
    VAR-\textit{d}24 & 1.0B & 5.25 & 6.21 \\
    VAR-\textit{d}30 & 2.0B & 4.96 & 5.92 \\
    \midrule
    RAR-B & 216M & 5.02 & 6.16 \\
    RAR-L & 461M & 4.73 & 5.78 \\
    RAR-XL & 955M & 4.66 & 5.45 \\
    RAR-XXL & 1499M & 4.50 & 5.21 \\
    \bottomrule
    \end{tabular}
\end{table}
\addtolength{\tabcolsep}{\mycolspace} 

\textbf{Robustness of \ours.} We analyze the robustness of \ours against a wide range of attacks, namely:
1) \textbf{Noise:} Adds Gaussian noise with a std of 0.1 to the image,
2) \textbf{Kernel:} Noise attack + application of a Gaussian blur with a kernel size of 7,
3) \textbf{Color:} Color jitter with a random hue of 0.3, saturation scaling of 3.0 and contrast of 3.0,
4) \textbf{Grey:} Transformation to grayscale, 
5) \textbf{JPEG:} JPEG compression to 75\% quality,
6) \textbf{SD-VAE:} Encoding and decoding of the images with the Stable Diffusion 2.1 VAE,
7) \textbf{CtrlRegen:} A regenerative attack based on the noising-denoising process in the latent space~\cite{liu2025image_ctrlregen}. \Cref{tab:tpr_attack_results} shows that \ours is robust against all these attacks.

\begin{table}[h!]
\setlength{\tabcolsep}{3pt}
\caption{\textbf{Robustness of \ours.} TPR@FPR=1\% for different attacks and RAR models, \ie 1\% corresponds to random guessing in watermark detection.}
\centering
\tiny
\begin{tabular}{lccccccccc}
\toprule
Model$\downarrow$ / Attacks$\rightarrow$ & None & Noise & Kernel & Color & Grey  & JPEG & SD-VAE & CtrlRegen\\
\midrule
VAR-\textit{d}16 & 99.30 & 45.32 & 42.80 & 56.20 & 88.10  & 88.20 & 54.00 & 4.70 \\
VAR-\textit{d}20 & 98.30 & 50.20 & 47.00 & 59.80 & 91.00  & 90.50 & 62.40 & 8.50 \\
VAR-\textit{d}24 & 99.30 & 51.10 & 49.10 & 60.30 & 90.50  & 90.40 & 64.40 & 8.40\\
VAR-\textit{d}30 & 99.00 & 49.20 & 45.80 & 57.10 & 87.60  & 87.80 & 60.20 & 6.90 \\
\midrule
RAR-B  & 96.50 & 16.58 & 14.04 & 12.73 & 38.20  & 81.57 & 85.33 & 16.20 &  \\
RAR-L  & 95.70 & 15.33 & 12.59 & 12.28 & 32.00  & 80.32 & 83.74 & 10.00 &  \\
RAR-XL & 96.60 & 18.95 & 16.67 & 13.95 & 36.70  & 85.28 & 87.58 & 17.80 &  \\
RAR-XXL & 93.80 & 14.44 & 12.86 & 11.52 & 25.60  & 78.05 & 79.81 & 4.40 &  \\
\bottomrule
\end{tabular}
\label{tab:tpr_attack_results}
\end{table}
\vspace{-1em}

%% file: content/07_conclusions.tex
\section{Conclusion}
We analyzed watermark radioactivity in image generative models and found that existing watermarking approaches for LDMs fail to retain radioactivity.
Motivated by this finding, we introduced \ours, the first watermarking method for IARs, designed to ensure radioactivity and enable provenance tracking even when generated images are used for training new IAR models. 
Through extensive experiments, we demonstrated the effectiveness of our approach in preserving watermark radioactivity, providing a robust solution for detecting unauthorized use of generated images.

%% file: content/08_app_background.tex
\section{Additional Related Work}

\subsection{Image Autoregressive Models (IARs)}

RARs (Randomized Autoregressive Models)~\citep{yu2024rar} randomly order the tokens and then train the model to correctly predict the next token in the sequence. Each token is assigned a position in the sequence, so the pretext task in RARs facilitates the multidirectional representation since each token can be predicted from any set of previously given other tokens. This reflects the nature of images where a given part of the image might be influenced by any not necessarily neighboring parts.
The randomization of token order in RARs is slowly reduced (annealed) from fully random order to perfect raster-scan order. 

\subsection{Watermarking Properties}

In text-to-image models, watermarks are designed to be imperceptible to the human eye yet detectable by a specialized detection algorithms. Image watermarks must possess the following essential properties: (1) \textit{Undetectability} ensures that the watermark remains imperceptible to unauthorized parties, (2) \textit{Unforgeability} guarantees that an adversary cannot reproduce the same watermark, (3) 
\textit{Tamper-evidence} allows for the detection of any modifications to the watermark, and (4) \textit{Robustness} ensures that the watermark remains detectable even under adversarial attacks. 

\section{Analysis of Radioactivity for DIAGNOSIS}

The DIAGNOSIS~\citep{wang2024diagnosis} method detects unauthorized data usage by injecting memorization into text-to-image DMs when they are trained on protected images. For this goal, the protected images are coated using the WaNet~\citep{nguyen2021wanet} injection function. A binary classifier is then trained on the coated and original images. When a DM $M_1$ is fine-tuned on the protected images, then the generated images, will be classifiable by the trained binary classifier to detect if the model has been trained with protected images.

\begin{figure}[htbp]
    \centering
    \begin{subfigure}{0.45\linewidth}
        \centering
        \includegraphics[width=\linewidth]{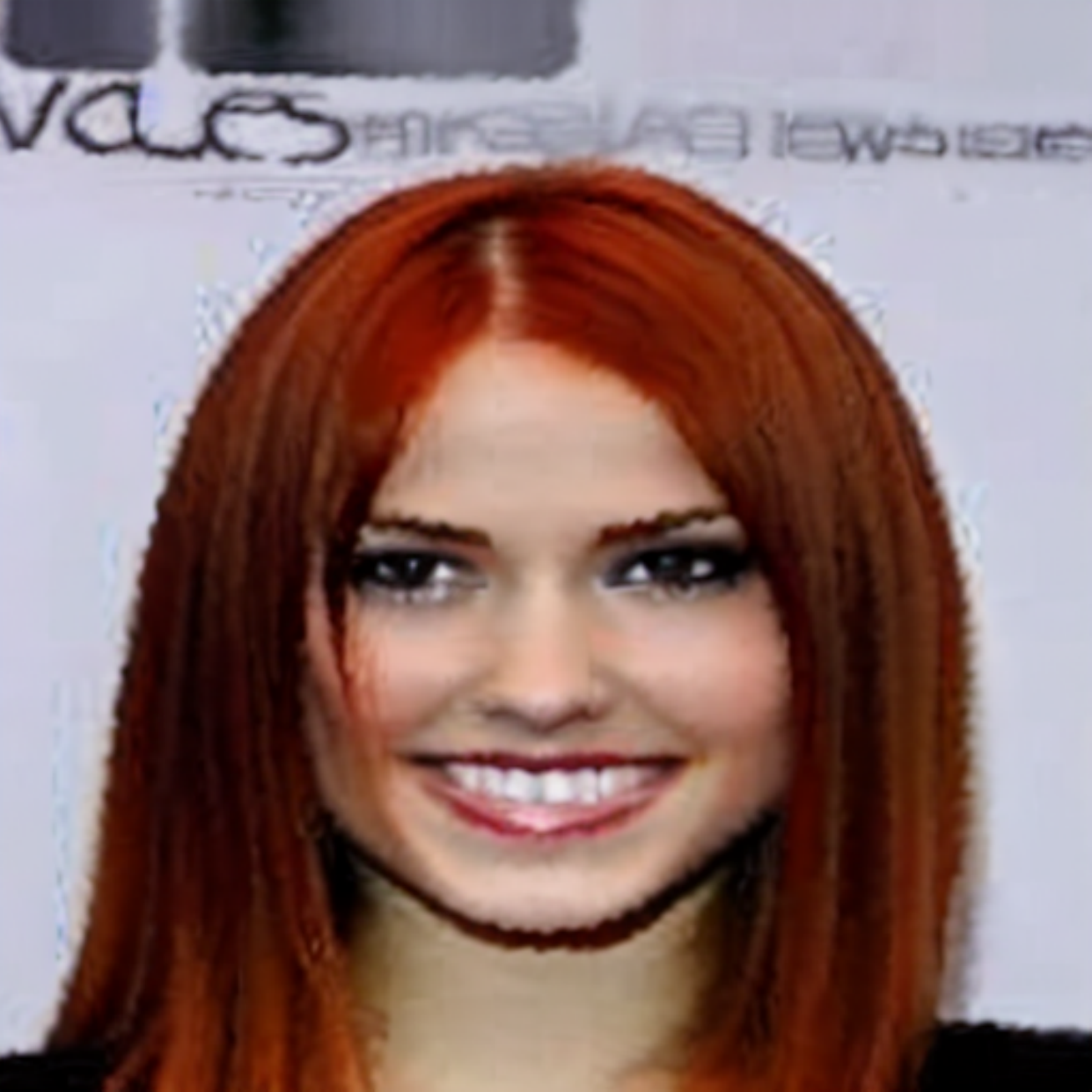}
        \caption{Image generated by DM $M_1$ after fine-tuning on the images coated with WaNet~\citep{nguyen2021wanet}.}
        \label{fig:DIAGNOSIS_sub1}
    \end{subfigure}
    \hfill
    \begin{subfigure}{0.45\linewidth}
        \centering
        \includegraphics[width=\linewidth]{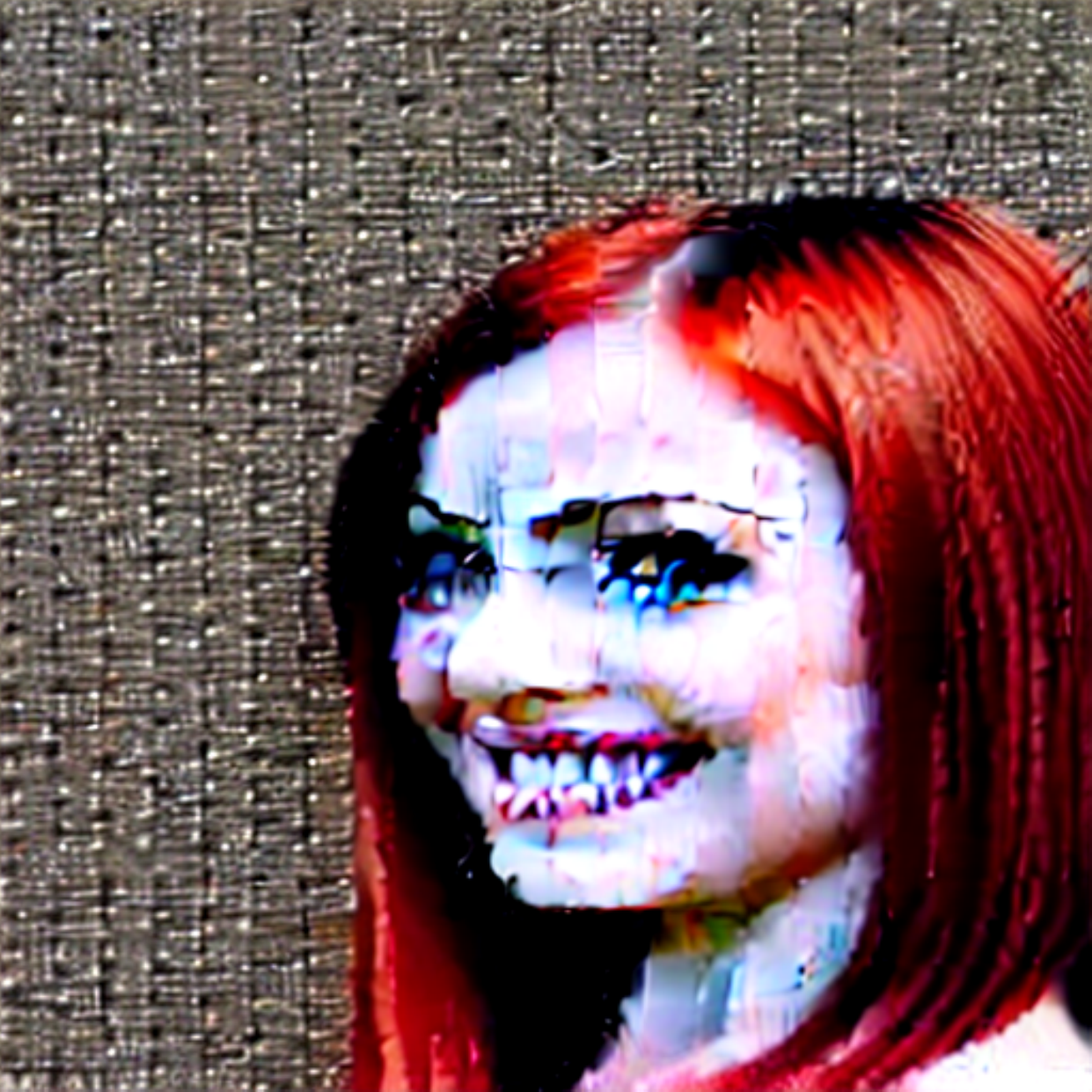}
        \caption{Image generated by DM $M_2$ after fine-tuning on images generated by DM $M_1$.}
        \label{fig:DIAGNOSIS_sub2}
    \end{subfigure}
    \vspace{-0.1cm}
    \caption{Comparison of the outputs from DM $M_1$ vs DM $M_2$ for the DIAGNOSIS protection method.}
    \label{fig:compare_DIAGNOSIS}
    \vspace{-0.5cm}
\end{figure}

To analyze if this method is radioactive, we used the code from~\citep{wang2024diagnosis} to train Stable Diffusion v1.4 for 100 epochs on 783 coated images from the CelebA~\citep{CelebA_dataset_deeplearningfaceattributes} dataset with unconditional triggering, a learning rate of 0.0001, a mini-batch of size 1 and using LoRA~\citep{hu2021lora} fine-tuning. 
After training the first DM $M_1$, we generate 783 images, with the prompts of the coated images. On these generated images, we evaluate the memorization strength using the binary classifier, which reliably classifies the model as malicious with a $100\%$ memorization strength. An example image can be seen in \cref{fig:DIAGNOSIS_sub1}. Then we train a second DM $M_2$ on the images generated by DM $M_1$, with the same hyperparameters, to analyze the transfer capabilities of the DIAGNOSIS method. Using the binary classifier on the images generated by DM $M_2$, results in a memorization strength of $75\%$, which is the threshold by which a model is classified as memorizing, meaning that the detection strength of the method degrades with each transfer. At the same time also the image quality of the DM is much worse, as can be seen in \cref{fig:DIAGNOSIS_sub2}. 

So while DIAGNOSIS displays capabilities in terms of transfer, it strongly impacts the generated image quality, making it impractical as a watermark substitution, which would make DMs \textit{radioactive}.

%% file: content/09_appendix.tex
\section{Experimental details}

The EDM model is trained unconditionally for 200 epochs with a mini-batch of size 512 with other hyperparameters following \cite{zhao2023recipe}. In our experiments for LDMs, we perform full finetuning of the Stable Diffusion 2.1~\citep{rombach2022high} model on the MSCOCO~\citep{lin2014microsoft} or its synthetic counterpart regenerated with the LDM. We finetune all models for 10 epochs, with batch size = $4$ and learning rate = $1e-4$. We use $512\times512$ images.
The RAR and VAR $M_2$ models are trained on 40000 generated watermarked ImageNet images for 10 epochs, with batch size = $4$, learning rate = $1e-4$ and a $256\times256$ resolution. The other model specific hyperparameters are taken from the original training setup as specified in \citep{yu2024rar} and \citep{tian2024visualVAR}. For a more consistent training we utilize only major-row scan in RAR for the whole fine-tuning process. For RAR we generate images with a $ \delta$ of 2.0 and for VAR with a $ \delta$ of 6.0. For both architectures we utilize a $\gamma$ of 0.25.

\section{Additional Results}
In \Cref{tab:after_ae} we analyze the effects of passing the watermarked images through the VAE. 

\renewcommand{\mycolspace}{2pt}
\addtolength{\tabcolsep}{-\mycolspace} 
\begin{table}[h!]
    \caption{\textbf{For \textit{Stable Signature}, \textit{PRC}, \textit{Tree-Ring}, and \textit{WIAR} methods the watermark remains intact after being encoded into the latent space of the SD 2.1.}  We report the bitwise accuracy (BA) for \textit{Stable Signature}, and \textit{PRC} (\ie 50\% corresponds to random guessing), as well as TPR@FPR=1\% for \textit{Tree-Ring} and \textit{\ours} (\ie 1\% corresponds to random guessing).}
    \label{tab:after_ae}
    \centering
    \tiny
    \begin{tabular}{cccccc}\toprule
    \textbf{Type of $M_1$} & \textbf{Type of $M_2$} & \textbf{Method} & \textbf{Metric} & \textbf{Output of $M_1$} & \textbf{After AE} \\
    \midrule
    LDM & LDM & Stable Signature & BA & $100.0$ & $100.0$ \\
    LDM & LDM & PRC & BA & $100.0$ & $100.0$ \\
    LDM & LDM & Tree-Ring & TPR@FPR=1\% & $100.0$  & $100.0$  \\
    RAR-B & LDM & \ours & TPR@FPR=1\% & $100.0$  & $100.0$  \\
    \bottomrule
    \end{tabular}
    \vspace{-1em}
\end{table}
\addtolength{\tabcolsep}{\mycolspace}

\begin{table}[h]
    \scriptsize
    \caption{\textbf{Wall-clock time} (in seconds) for an image generation on NVIDIA A40 GPU without and with \ours, which adds a relatively small overhead, similarly to LLM watermarking.}
    \centering
    \resizebox{1.0\linewidth}{!}{%
    \begin{tabular}{c|cccc|cccc}
    \toprule
    Setup & RAR-B & RAR-L & RAR-XL & RAR-XXL & VAR 16 & VAR 20 & VAR 24 & VAR 30\\
    \midrule
    original & 4.043 & 4.129 & 5.430 & 6.623 & 0.399 & 0.428 & 0.462 & 0.531 \\
    \ours & 6.781 & 6.889 & 8.201 & 9.411 & 0.566 & 0.587 & 0.605 & 0.641 \\
    \bottomrule
    \end{tabular}
    }
    \label{tab:time}
\end{table}

\noindent \textbf{Post and pre-processing watermarks.}
We analyze the watermarks Recipe, RivaGAN, ZoDiac for IARs in \Cref{tab:additional_watermarks} and found that VQ-VAE tokenization in VAR and RAR removes them. As an effect, Recipe---a pre-processing watermark---does not embed into IARs during training. RivaGAN and ZoDiac, as post-processing watermarks, can mark generated images but do not transfer from model $M_1$ to $M_2$ one.

\begin{table}[h]
    \centering
    \scriptsize
    \caption{We report the TPR@1\%FPR for different post and pre-processing watermarks after encoding with the RAR or VAR VQ-VAE.}
    \setlength{\tabcolsep}{3pt}
    \begin{tabular}{ccccc}
        \toprule
        \textbf{Method} & \textbf{Watermarked Data} & \textbf{After VQ-VAE (RAR)} & \textbf{After VQ-VAE (VAR)} \\
        \midrule
        Recipe   & $100$ & $0.5$ & $0.8$ \\
        RivaGAN  & $100$ & $0.9$ & $1.0$ \\
        Zodiac  & $100$ & $0.9$ & $1.0$ \\
        \bottomrule
    \end{tabular}
    \label{tab:additional_watermarks}
\end{table}

\textbf{Watermark transferability.} We perform additional experiments of the transferability of \ours for different model sizes and cross-modality below in \Cref{tab:transferability}. The results show that \ours is radioactive in the same modality setting for different sizes (RAR to RAR and VAR to VAR). However, WIAR does not transfer cross-architecture (RAR to VAR \& vice versa), see last two rows below, since different encoders are used in RAR \& VAR, which can be patched by using a single best as in LDMs. 

\begin{table}[h]
    \scriptsize
    \centering
    \caption{We follow the setting in \Cref{tab:watermark_single_tab} and enforce transfer of a single watermarked image.}
    \begin{tabular}{cccccc}\toprule
    \textbf{Type of $M_1$} & \textbf{Type of $M_2$}&\textbf{Output of $M_1$} & \textbf{Output of $M_2$} \\
    \midrule
    RAR-B & RAR-XXL &$100$ & $ 100 $\\
    RAR-L & RAR-B & $100$ & $100$ \\
    RAR-XL & RAR-B & $100$ & $100$ \\
    RAR-XXL & RAR-B &$100$ & $93.5$\\
    \midrule
    VAR 16 & VAR 30 &$100$ & $ 100 $\\
    VAR 20 & VAR 16  & $100$ & $100$ \\
    VAR 24 & VAR 16 & $100$ & $100$ \\
    VAE 30 & VAR 16  &$100$ & $95.2$\\
    \midrule
    RAR-B  & VAR 16 & $100$ & $1$ \\
    VAR 16 & RAR-B & $100$ & $1$ \\ 
    \bottomrule
    \end{tabular}
    \label{tab:transferability}
\end{table}

\textbf{Image Quality vs Watermark Detection.} We analyze the trade-off between watermark strength (TPR@FPR=1\%) and image quality (FID) computed for 1000 images at different watermarking strength levels (bias $\delta$ dependent) in \Cref{tab:watermark_strength}. With higher $\delta$ the image quality degrades gracefully, the detection of \ours rapidly increases ($\delta=0$ is no watermark).

\begin{table}[h!]
    \centering
    \scriptsize
    \caption{\textbf{\ours shows a graceful tradeoff between watermarking strength and image quality.} We compute the FID and TPR@1\%FPR for 1.000 images against different delta values.}
    \setlength{\tabcolsep}{2pt}
    \resizebox{\columnwidth}{!}{%
    \begin{tabular}{lcccccc|ccccc}
    \toprule
    & \multicolumn{6}{c|}{FID} & \multicolumn{5}{c}{TPR@FPR=1\%} \\
    Model & $\delta=0$ & $\delta=1$ & $\delta=2$ & $\delta=3$ & $\delta=4$ & $\delta=5$ & $\delta=1$ & $\delta=2$ & $\delta=3$ & $\delta=4$ & $\delta=5$ \\
    \midrule
    RAR-B   & 38.3 & 38.35 & 39.0 & 40.4 & 40.9 & 42.3 & 50.2 & 96.5 & 99.4 & 100.0 & 100.0 \\
    RAR-L   & 38.0 & 38.54 & 38.8 & 39.9 & 40.9 & 41.1 & 48.7 & 95.7 & 99.7 & 99.8 & 99.9 \\
    RAR-XL  & 38.2 & 38.31 & 38.9 & 39.1 & 39.6 & 40.1 & 50.8 & 96.6 & 99.1 & 99.8 & 99.8 \\
    RAR-XXL & 38.0 & 38.06 & 38.1 & 39.1 & 39.0 & 39.5 & 41.6 & 93.8 & 97.9 & 99.7 & 99.8 \\
    \bottomrule
    \end{tabular}
    }

    \vspace{0.5em}

    \resizebox{\columnwidth}{!}{%
    \begin{tabular}{lcccccc|ccccc}
    \toprule
    & \multicolumn{6}{c|}{FID} & \multicolumn{5}{c}{TPR@FPR=1\%} \\
    Model & $\delta=0$ & $\delta=1$ & $\delta=2$ & $\delta=4$ & $\delta=6$ & $\delta=8$ & $\delta=1$ & $\delta=2$ & $\delta=4$ & $\delta=6$ & $\delta=8$ \\
    \midrule
    VAR-16 & 45.9 & 46.1 & 47.9 & 48.2 & 48.4 & 49.1 & 29.7 & 89.0 & 98.3 & 99.3 & 99.3 \\
    VAR-20 & 43.5 & 44.1 & 45.6 & 46.3 & 46.6 & 47.1 & 34.7 & 91.3 & 99.7 & 98.3 & 99.7 \\
    VAR-24 & 43.3 & 43.9 & 45.3 & 45.9 & 46.5 & 46.9 & 53.6 & 95.3 & 99.0 & 99.2 & 99.3 \\
    VAR-30 & 42.3 & 43.0 & 43.6 & 44.2 & 44.6 & 45.1 & 56.7 & 94.0 & 98.3 & 99.0 & 99.3 \\
    \bottomrule
    \end{tabular}
    }
    \label{tab:watermark_strength}
\end{table}

\textbf{Image quality in $M_2$.} \Cref{tab:image_quality_transfer} shows the FID-scores for different model sizes and stages on 1000 images. $M_2$ is fine-tuned on generated watermarked data, which slightly decreases the performance, compared to training on natural data as in $M_1$, but also in this case, our \ours is radioactive. 

\begin{table}[h]
    \centering
    \scriptsize
    \caption{\textbf{The radioactivity of \ours has limited impact on M2 image quality.} We compute the FID for 1.000 images generated by M1 and M2.}
    \resizebox{1.0\linewidth}{!}{%
    \begin{tabular}{ccccc|cccc}
        \toprule
         Model &  RAR-B & RAR-L & RAR-XL & RAR-XXL & VAR 16 & VAR 20 & VAR 24 & VAR 30 \\
         \midrule
        $M_1$ & 39.0 & 38.8 & 38.9 & 38.1 & 45.9 & 43.5 & 43.3 & 42.3 \\
        $M_2$ & 47.5 & 40.1 & 44.5 & 43.4 & 56.2  & 49.7 & 48.2 & 47.0 \\
        \bottomrule
    \end{tabular}
    }
    \label{tab:image_quality_transfer}
\end{table}

\section{Additional Figures}

We present a qualitative comparison between the outputs from $M_1$ and $M_2$ in \Cref{fig:app_compare_outputs_M1_M2}. 

\begin{figure}[htbp]
    \centering
    \begin{subfigure}{0.45\textwidth}
        \centering
        \includegraphics[width=0.75\linewidth]{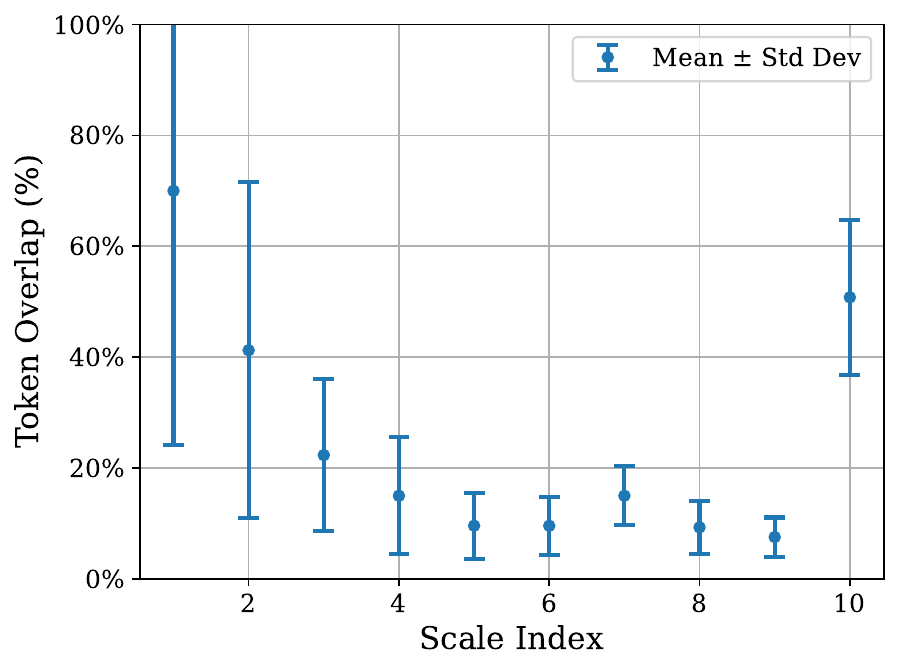}
        \caption{Token overlap for the VAR model with depth 16.}
    \end{subfigure}
    \hfill
    \begin{subfigure}{0.45\textwidth}
        \centering
        \includegraphics[width=0.75\linewidth]{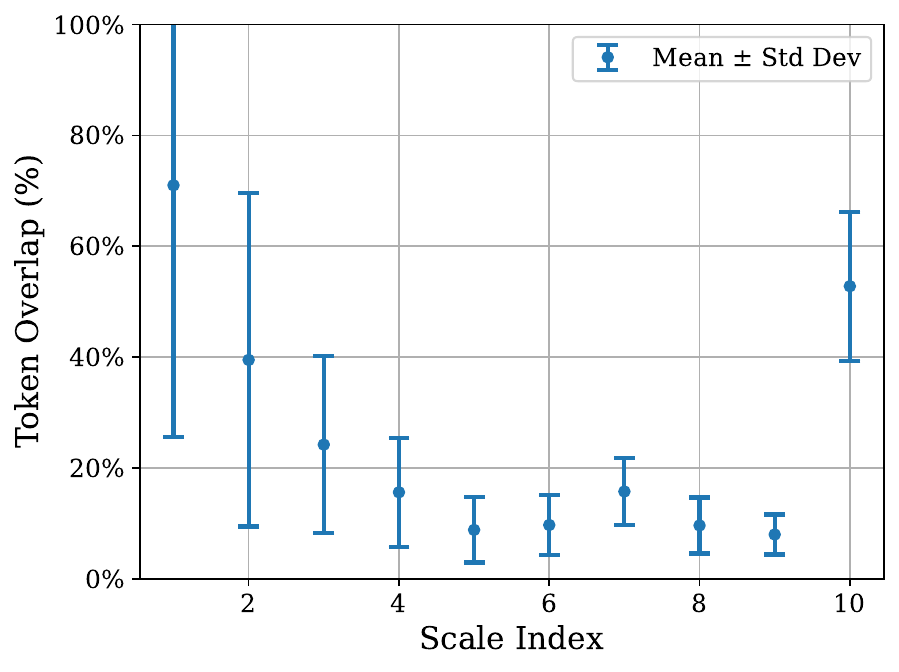}
        \caption{Token overlap for the VAR model with depth 20.}
    \end{subfigure}
    \hfill
    \begin{subfigure}{0.45\textwidth}
        \centering
        \includegraphics[width=0.75\linewidth]{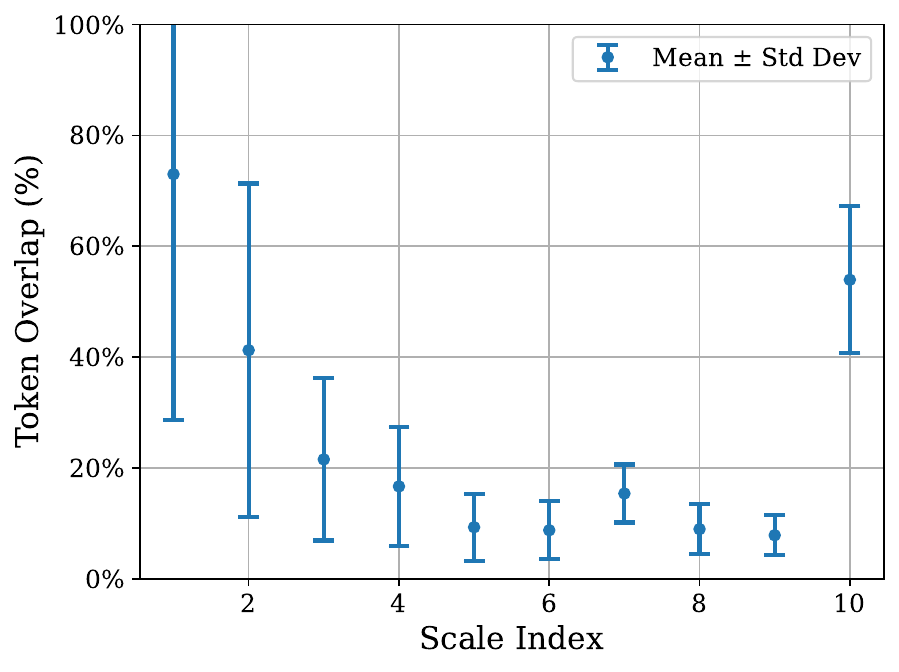}
        \caption{Token overlap for the VAR model with depth 24.}
    \end{subfigure}
    \hfill
    \begin{subfigure}{0.45\textwidth}
        \centering
        \includegraphics[width=0.75\linewidth]{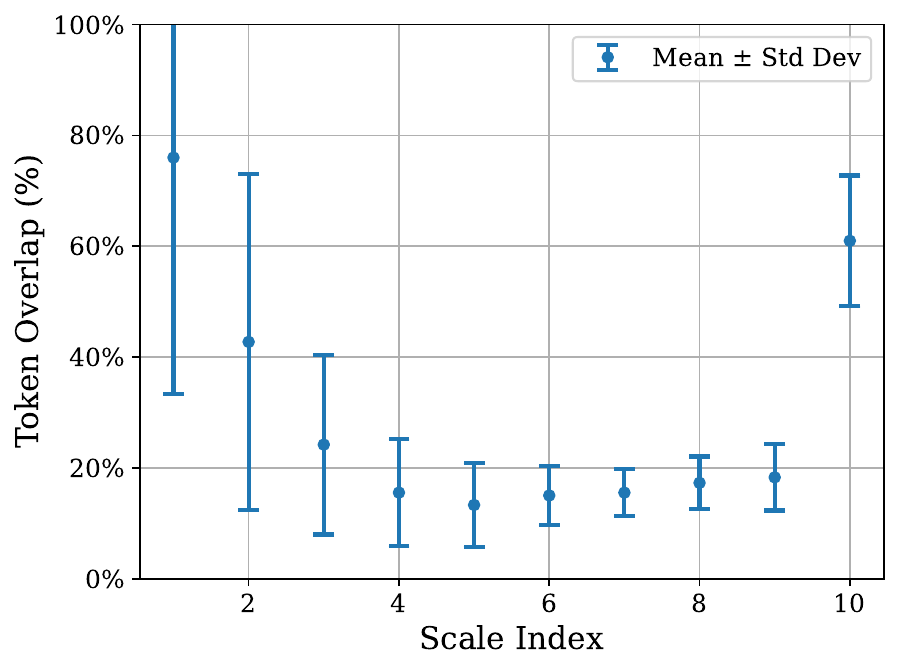}
        \caption{Token overlap for the VAR model with depth 30.}
    \end{subfigure}
    \caption{\textbf{Comparison of the token overlap at the different scales (resolutions) and for different depths (16, 20, 24, and 36) of the VAR model.}}
    \label{fig:compare_token_overlap}
\end{figure}

The comparison of token overlap at the different scales (resolutions) and for different depths (16, 20, 24, and 36) of the VAR model are shown in \Cref{fig:compare_token_overlap}.

